# Transductive Rademacher Complexity and its Applications

**Ran El-Yaniv**                         RANI@CS.TECHNION.AC.IL
**Dmitry Pechyony**             PECHYONY@CS.TECHNION.AC.IL
*Department of Computer Science*
*Technion - Israel Institute of Technology*
*Haifa, 32000, Israel*

## Abstract

We develop a technique for deriving data-dependent error bounds for transductive learning algorithms based on transductive Rademacher complexity. Our technique is based on a novel general error bound for transduction in terms of transductive Rademacher complexity, together with a novel bounding technique for Rademacher averages for particular algorithms, in terms of their "unlabeled-labeled" representation. This technique is relevant to many advanced graph-based transductive algorithms and we demonstrate its effectiveness by deriving error bounds to three well known algorithms. Finally, we present a new PAC-Bayesian bound for mixtures of transductive algorithms based on our Rademacher bounds.

## 1. Introduction

Alternative learning models that utilize unlabeled data have received considerable attention in the past few years. Two prominent models are semi-supervised and transductive learning. The main attraction of these models is theoretical and empirical evidence (Chapelle, Schölkopf, & Zien, 2006) indicating that they can often allow for more efficient and significantly faster learning in terms of sample complexity. In this paper we support the theoretical evidence by providing risk bounds for a number of state-of-the-art transductive algorithms. These bounds utilize both labeled and unlabeled examples and can be much tighter than the bounds relying on labeled examples alone.

Here we focus on distribution-free transductive learning. In this setting we are given a labeled training sample as well as an unlabeled test sample. The goal is to guess the labels of the *given* test points as accurately as possible[1]. Rather than generating a general hypothesis capable of predicting the label of any point, as in inductive learning, it is advocated by Vapnik (1982) that we should aim in transduction to solve an easier problem by transferring knowledge directly from the labeled points to the unlabeled ones.

Transductive learning was already proposed and briefly studied more than thirty years ago by Vapnik and Chervonenkis (1974), but only lately has it been empirically recognized that transduction can often facilitate more efficient or accurate learning than the traditional supervised learning approach (Chapelle et al., 2006). This recognition has motivated a flurry of recent activity focusing on transductive learning, with many new algorithms and heuristics being proposed. Nevertheless, issues such as the identification of "universally" effective learning principles for transduction remain unresolved. Statistical learning theory provides a principled approach for attacking such questions through the study of error bounds. For example, in inductive learning such bounds have proven instrumental in characterizing learning principles and deriving practical algorithms (Vapnik, 2000).

---

1. Many papers refer to this model as semi-supervised learning. However, the setting of semi-supervised learning is different from transduction. In semi-supervised learning the learner is given randomly drawn training set consisting of labeled and unlabeled examples. The goal of the learner is to generate a hypothesis providing accurate predictions on the *unseen* examples.

In this paper we consider the classification setting of transductive learning. So far, several general error bounds for transductive classification have been developed by Vapnik (1982), Blum and Langford (2003), Derbeko, El-Yaniv, and Meir (2004), El-Yaniv and Pechony (2006). We continue this fruitful line of research and develop a new technique for deriving explicit data-dependent error bounds. These bounds are less tight than implicit ones, developed by Vapnik and by Blum and Langford. However the explicit bounds may potentially be used for model selection and guide the development of new learning algorithms.

Our technique consists of two parts. In the first part we develop a novel general error bound for transduction in terms of transductive Rademacher complexity. While this bound is syntactically similar to known inductive Rademacher bounds (see, e.g., Bartlett & Mendelson, 2002), it is fundamentally different in the sense that the transductive Rademacher complexity is computed with respect to the hypothesis space that can be chosen *after* observing unlabeled training and test examples. This opportunity is unavailable in the inductive setting where the hypothesis space must be fixed *before* any example is observed.

The second part of our bounding technique is a generic method for bounding Rademacher complexity of transductive algorithms based on their *unlabeled-labeled representation (ULR)*. In this representation the soft-classification vector generated by the algorithm is a product $U\boldsymbol{\alpha}$, where $U$ is a matrix that depends on the unlabeled data and $\boldsymbol{\alpha}$ is a vector that may depend on all given information, including the labeled training set. Any transductive algorithm has infinite number of ULRs, including a trivial ULR, with $U$ being an identity matrix. We show that many state-of-the-art algorithms have non-trivial ULR leading to non-trivial error bounds. Based on ULR representation we bound Rademacher complexity of transductive algorithms in terms of the spectrum of the matrix $U$ in their ULR. This bound justifies the spectral transformations, developed by Chapelle, Weston, and Schölkopf (2003), Joachims (2003), Johnson and Zhang (2008), that are commonly done to improve the performance of transductive algorithms. We instantiate the Rademacher complexity bound for the "consistency method" of Zhou et al. (2004), the spectral graph transducer (SGT) algorithm of Joachims (2003) and the Tikhonov regularization algorithm of Belkin, Matveeva, and Niyogi (2004). The bounds obtained for these algorithms are explicit and can be easily computed.

We also show a simple Monte-Carlo scheme for bounding the Rademacher complexity of any transductive algorithm using its ULR. We demonstrate the efficacy of this scheme for the "consistency method" of Zhou et al. (2004). Our final contribution is a PAC-Bayesian bound for transductive mixture algorithms. This result, which is stated in Theorem 4, is obtained as a consequence of Theorem 2 using the techniques of Meir and Zhang (2003). This result motivates the use of ensemble methods in transduction that are yet to be explored in this setting.

The paper has the following structure. In Section 1.1 we survey the results that are closely related to our work. In Section 2 we define our learning model and transductive

Rademacher complexity. In Section 3 we develop a novel concentration inequality for functions over partitions of the finite set of points. This inequality and transductive Rademacher complexity are used in Section 4 to derive uniform risk bound, which depends on transductive Rademacher complexity. In Section 5 we introduce a generic method for bounding Rademacher complexity of any transductive algorithm using its unlabeled-labeled representation. In Section 6 we exemplify this technique to obtain explicit risk bounds for several known transductive algorithms. Finally, in Section 7 we instantiate our risk bound to transductive mixture algorithms. We discuss directions for future research in Section 8. The technical proofs of our results are presented in Appendices A-I.

Preliminary (and shorter) version of this paper has appeared in the Proceedings of the 20th Annual Conference on Learning Theory, page 157–171, 2007.

## 1.1 Related Work

Vapnik (1982) presented the first general 0/1 loss bounds for transductive classification. His bounds are implicit in the sense that tail probabilities are specified in the bound as the outcome of a computational routine. Vapnik's bounds can be refined to include prior "beliefs" as noted by Derbeko et al. (2004). Similar implicit but somewhat tighter bounds were developed by Blum and Langford (2003) for the 0/1 loss case. Explicit PAC-Bayesian transductive bounds for any bounded loss function were presented by Derbeko et al. (2004). Catoni (2004, 2007) and Audibert (2004) developed PAC-Bayesian and VC dimension-based risk bounds for the special case when the size of the test set is a multiple of the size of the training set. Unlike our PAC-Bayesian bound, the published transductive PAC-Bayesian bounds hold for deterministic hypotheses and for Gibbs classifiers. The bounds of Balcan and Blum (2006) for semi-supervised learning also hold in the transductive setting, making them conceptually similar to some transductive PAC-Bayesian bounds. General error bounds based on stability were developed by El-Yaniv and Pechyony (2006).

Effective applications of the general bounds mentioned above to particular algorithms or "learning principles" is not automatic. In the case of the PAC-Bayesian bounds several such successful applications were presented in terms of appropriate "priors" that promote various structural properties of the data (see, e.g., Derbeko et al., 2004; El-Yaniv & Gerzon, 2005; Hanneke, 2006). Ad-hoc bounds for particular algorithms were developed by Belkin et al. (2004) and by Johnson and Zhang (2007). Unlike other bounds (including ours) the bound of Johnson and Zhang does not depend on the empirical error but only on the properties of the hypothesis space. If the size of the training and test set increases then their bound converges to zero[2]. Thus the bound of Johnson and Zhang effectively proves the consistency of transductive algorithms that they consider. However this bound holds only if the hyperparameters of those algorithms are chosen w.r.t. to the unknown test labels. Hence the bound of Johnson and Zhang cannot be computed explicitly.

Error bounds based on Rademacher complexity were introduced by Koltchinskii (2001) and are a well-established topic in induction (see Bartlett & Mendelson, 2002, and references therein). The first Rademacher transductive risk bound was presented by Lanckriet et al. (2004, Theorem 24). This bound, which is a straightforward extension of the inductive

---

2. in all other known explicit bounds the increase of training and test sets decreases only the slack term but not the empirical error.

Rademacher techniques of Bartlett and Mendelson (2002), is limited to the special case when training and test sets are of equal size. The bound presented here overcomes this limitation.

## 2. Definitions

In Section 2.1 we provide a formal definition of our learning model. Then in Section 2.2 we define transductive Rademacher complexity and compare it with its inductive counterpart.

### 2.1 Learning Model

In this paper we use a distribution-free transductive model, as defined by Vapnik (1982, Section 10.1, Setting 1). Consider a fixed set $S_{m+u} \triangleq \{(x_i, y_i)\}_{i=1}^{m+u}$ of $m + u$ points $x_i$ in some space together with their labels $y_i$. The learner is provided with the (unlabeled) *full-sample* $X_{m+u} \triangleq \{x_i\}_{i=1}^{m+u}$. A set consisting of $m$ points is selected from $X_{m+u}$ uniformly at random among all subsets of size $m$. These $m$ points together with their labels are given to the learner as a *training set*. Re-numbering the points we denote the unlabeled training set points by $X_m \triangleq \{x_1, \ldots, x_m\}$ and the labeled training set by $S_m \triangleq \{(x_i, y_i)\}_{i=1}^{m}$. The set of unlabeled points $X_u \triangleq \{x_{m+1}, \ldots, x_{m+u}\} = X_{m+u} \setminus X_m$ is called the *test set*. The learner's goal is to predict the labels of the test points in $X_u$ based on $S_m \cup X_u$.

**Remark 1** *In our learner model each example $x_i$ has unique label $y_i$. However we allow that for $i \neq j$, $x_i = x_j$ but $y_i \neq y_j$.*

The choice of the set of $m$ points as described above can be viewed in three equivalent ways:

1. Drawing $m$ points from $X_{m+u}$ uniformly *without replacement*. Due to this draw, the points in the training and test sets are *dependent*.

2. Random permutation of the full sample $X_{m+u}$ and choosing the first $m$ points as a training set.

3. Random partitioning of $m + u$ points into two disjoint sets of $m$ and $u$ points.

To emphasize different aspects of the transductive learning model, throughout the paper we use interchangeably these three views on the generation of the training and test sets.

This paper focuses on binary learning problems where labels $y \in \{\pm 1\}$. The learning algorithms we consider generate "soft classification" vectors $\mathbf{h} = (h(1), \ldots h(m + u)) \in \mathbb{R}^{m+u}$, where $h(i)$ (or $h(x_i)$) is the soft, or confidence-rated, label of example $x_i$ given by the "hypothesis" $\mathbf{h}$. For actual (binary) classification of $x_i$ the algorithm outputs $\mathrm{sgn}(h(i))$. We denote by $\mathcal{H}_{\mathrm{out}} \subseteq \mathbb{R}^{m+u}$ the set of all possible soft classification vectors (over all possible training/test partitions) that are generated by the algorithm.

Based on the full-sample $X_{m+u}$, the algorithm selects an hypothesis space $\mathcal{H} \subseteq \mathbb{R}^{m+u}$ of soft classification hypotheses. Note that $\mathcal{H}_{\mathrm{out}} \subseteq \mathcal{H}$. Then, given the labels of training points the algorithm outputs one hypothesis $\mathbf{h}$ from $\mathcal{H}_{\mathrm{out}} \cap \mathcal{H}$ for classification. The goal of the transductive learner is to find a hypothesis $\mathbf{h}$ minimizing the *test error* $\mathcal{L}_u(\mathbf{h}) \triangleq$ $\frac{1}{u} \sum_{i=m+1}^{m+u} \ell(h(i), y_i)$ w.r.t. the 0/1 loss function $\ell$. The *empirical error* of $\mathbf{h}$ is $\widehat{\mathcal{L}}_m(\mathbf{h}) \triangleq \frac{1}{m} \sum_{i=1}^{m} \ell(h(i), y_i)$ and the *full sample error* of $\mathbf{h}$ is $\mathcal{L}_{m+u}(\mathbf{h}) \triangleq \frac{1}{m+u} \sum_{i=1}^{m+u} \ell(h(i), y_i)$. In this work we also use the margin loss function $\ell_\gamma$. For a positive real $\gamma$, $\ell_\gamma(y_1, y_2) = 0$ if $y_1 y_2 \geq \gamma$ and $\ell_\gamma(y_1, y_2) = \min\{1, \ 1 - y_1 y_2/\gamma\}$ otherwise. The *empirical (margin) error* of $\mathbf{h}$ is $\widehat{\mathcal{L}}_m^\gamma(\mathbf{h}) \triangleq \frac{1}{m} \sum_{i=1}^{m} \ell_\gamma(h(i), y_i)$. We denote by $\mathcal{L}_u^\gamma(\mathbf{h})$ the margin error of the test set and by $\mathcal{L}_{m+u}^\gamma(\mathbf{h})$ the margin full sample error.

We denote by $I_r^s$, $r < s$, the set of natural numbers $\{r, r+1, \ldots, s\}$. Throughout the paper we assume that the vectors are column ones. We mark all vectors with the boldface.

## 2.2 Transductive Rademacher Complexity

We adapt the inductive Rademacher complexity to our transductive setting but generalize it a bit to also include "neutral" Rademacher values.

**Definition 1 (Transductive Rademacher complexity)** *Let $\mathcal{V} \subseteq \mathbb{R}^{m+u}$ and $p \in [0, 1/2]$. Let $\boldsymbol{\sigma} = (\sigma_1, \ldots, \sigma_{m+u})^T$ be a vector of i.i.d. random variables such that*

$$\sigma_i \triangleq \begin{cases} 1 & \text{with probability} \quad p; \\ -1 & \text{with probability} \quad p; \\ 0 & \text{with probability} \quad 1 - 2p. \end{cases} \tag{1}$$

*The* transductive Rademacher complexity *with parameter $p$ is*

$$R_{m+u}(\mathcal{V}, p) \triangleq \left( \frac{1}{m} + \frac{1}{u} \right) \cdot \mathbf{E}_{\boldsymbol{\sigma}} \left\{ \sup_{\mathbf{v} \in \mathcal{V}} \boldsymbol{\sigma}^T \cdot \mathbf{v} \right\} .$$

The need for this novel definition of Rademacher complexity is technical. Two main issues that lead to the new definition are:

1. The need to bound the test error $\mathcal{L}_u(\mathbf{h}) = \frac{1}{u} \sum_{i=m+1}^{m+u} \ell(h(i), y_i)$. Notice that in inductive risk bounds the standard definition of Rademacher complexity (see Definition 2 below), with binary values of $\sigma_i$, is used to bound the generalization error, which is an inductive analogue of the full sample error $\mathcal{L}_{m+u}(\mathbf{h}) = \frac{1}{m+u} \sum_{i=1}^{m+u} \ell(h(i), y_i)$.

2. Different sizes ($m$ and $u$ respectively) of training and test set.

See Section 4.1 for more technical details that lead to the above definition of Rademacher complexity.

For the sake of comparison we also state the inductive definition of Rademacher complexity.

**Definition 2 (Inductive Rademacher complexity, Koltchinskii, 2001)** *Let $\mathcal{D}$ be a probability distribution over $\mathcal{X}$. Suppose that the examples $X_n = \{x_i\}_{i=1}^{n}$ are sampled independently from $\mathcal{X}$ according to $\mathcal{D}$. Let $\mathcal{F}$ be a class of functions mapping $\mathcal{X}$ to $\mathbb{R}$. Let $\boldsymbol{\sigma} = \{\sigma_i\}_{i=1}^{n}$ be an independent uniform $\{\pm 1\}$-valued random variables, $\sigma_i = 1$ with probability $1/2$ and $\sigma_i = -1$ with the same probability. The* empirical Rademacher complex-ity is[3] $\widehat{R}_n^{(\text{ind})}(\mathcal{F}) \triangleq \frac{2}{n} \mathbf{E}_{\boldsymbol{\sigma}} \left\{ \sup_{f \in \mathcal{F}} \sum_{i=1}^n \sigma_i f(x_i) \right\}$ and the Rademacher complexity of $\mathcal{F}$ is $R_n^{(\text{ind})}(\mathcal{F}) \triangleq \mathbf{E}_{X_n \sim \mathcal{D}^n} \left\{ \widehat{R}_n^{(\text{ind})}(\mathcal{F}) \right\}$.

For the case $p = 1/2$, $m = u$ and $n \triangleq m + u$ we have that $R_{m+u}(\mathcal{V}) = 2\widehat{R}_{m+u}^{(\text{ind})}(\mathcal{V})$. Whenever $p < 1/2$, some Rademacher variables will attain (neutral) zero values and reduce the complexity (see Lemma 1). We use this property to tighten our bounds.

Notice that the transductive complexity is an empirical quantity that does not depend on any underlying distribution, including the one over the choices of the training set. Since in distribution-free transductive model the unlabeled full sample of training and test points is fixed, in transductive Rademacher complexity we don't need the outer expectation, which appears in the inductive definition. Also, the transductive complexity depends on both the (unlabeled) training and test points whereas the inductive complexity only depends only on the (unlabeled) training points.

The following lemma, whose proof appears in Appendix A, states that $R_{m+u}(\mathcal{V}, p)$ is monotone increasing with $p$. The proof is based on the technique used in the proof of Lemma 5 in the paper of Meir and Zhang (2003).

**Lemma 1** *For any $\mathcal{V} \subseteq \mathbb{R}^{m+u}$ and $0 \leq p_1 < p_2 \leq 1/2$, $R_{m+u}(\mathcal{V}, p_1) < R_{m+u}(\mathcal{V}, p_2)$.*

In the forthcoming results we utilize the transductive Rademacher complexity with $p_0 \triangleq \frac{mu}{(m+u)^2}$. We abbreviate $R_{m+u}(\mathcal{V}) \triangleq R_{m+u}(\mathcal{V}, p_0)$. By Lemma 1, all our bounds also apply to $R_{m+u}(\mathcal{V}, p)$ for all $p > p_0$. Since $p_0 < \frac{1}{2}$, the Rademacher complexity involved in our results is strictly smaller than the standard inductive Rademacher complexity defined over $X_{m+u}$. Also, if transduction approaches the induction, namely $m$ is fixed and $u \to \infty$, then $\widehat{R}_{m+u}^{(\text{ind})}(\mathcal{V}) \to 2R_{m+u}(\mathcal{V})$.

## 3. Concentration Inequalities for Functions over Partitions

In this section we develop a novel concentration inequality for functions over partitions and compare it to the several known ones. Our concentration inequality is utilized in the derivation of the forthcoming risk bound.

Let $\mathbf{Z} \triangleq \mathbf{Z}_1^{m+u} \triangleq (Z_1, \ldots, Z_{m+u})$ be a *random permutation vector* where the variable $Z_k$, $k \in I_1^{m+u}$, is the $k$th component of a permutation of $I_1^{m+u}$ that is chosen uniformly at random. Let $\mathbf{Z}^{ij}$ be a perturbed permutation vector obtained by exchanging the values of $Z_i$ and $Z_j$ in $\mathbf{Z}$. Any function $f$ on permutations of $I_1^{m+u}$ is called $(m, u)$-*permutation symmetric* if $f(\mathbf{Z}) \triangleq f(Z_1, \ldots, Z_{m+u})$ is symmetric on $Z_1, \ldots, Z_m$ as well as on $Z_{m+1}, \ldots, Z_{m+u}$.

In this section we present a novel concentration inequality for $(m, u)$-permutation symmetric functions. Note that an $(m, u)$-permutation symmetric function is essentially a function over the partition of $m + u$ items into sets of sizes $m$ and $u$. Thus, the forthcoming inequalities of Lemmas 2 and 3, while being stated for $(m, u)$-permutation symmetric functions, also hold in exactly the same form for functions over partitions. Conceptually it is more convenient to view our results as concentration inequalities for functions over partitions. However, from a technical point of view we find it more convenient to consider $(m, u)$-permutation symmetric functions.

---

3. The original definition of Rademacher complexity, as given by Koltchinskii (2001), is slightly different from the one presented here, and contains $\sup_{f \in \mathcal{F}} \left| \sum_{i=1}^n \sigma_i f(x_i) \right|$ instead of $\sup_{f \in \mathcal{F}} \sum_{i=1}^n \sigma_i f(x_i)$. However, from the conceptual point of view, Definition 2 and the one given by Koltchinskii are equivalent.

The following lemma (that will be utilized in the proof of Theorem 1) presents a concentration inequality that is an extension of Lemma 2 of El-Yaniv and Pechyony (2006). The proof (appearing in Appendix B) relies on McDiarmid's inequality (McDiarmid, 1989, Corollary 6.10) for martingales.

**Lemma 2** *Let $\mathbf{Z}$ be a random permutation vector over $I_1^{m+u}$. Let $f(\mathbf{Z})$ be an $(m, u)$-permutation symmetric function satisfying $\left| f(\mathbf{Z}) - f(\mathbf{Z}^{ij}) \right| \leq \beta$ for all $i \in I_1^m$, $j \in I_{m+1}^{m+u}$. Then*

$$\mathbf{P}_{\mathbf{Z}} \left\{ f(\mathbf{Z}) - \mathbf{E}_{\mathbf{Z}} \left\{ f(\mathbf{Z}) \right\} \geq \epsilon \right\} \leq \exp \left( -\frac{2\epsilon^2(m + u - 1/2)}{mu\beta^2} \left( 1 - \frac{1}{2\max(m, u)} \right) \right) \quad . \qquad (2)$$

The right hand side of (2) is approximately $\exp \left( -\frac{2\epsilon^2}{\beta^2} \left( \frac{1}{m} + \frac{1}{u} \right) \right)$. A similar, but less tight inequality can be obtained by reduction of the draw of random permutation to the draw of $\min(m, u)$ independent random variables and application of the bounded difference inequality of McDiarmid (1989):

**Lemma 3** *Suppose that the conditions of Lemma 2 hold. Then*

$$\mathbf{P}_{\mathbf{Z}} \left\{ f(\mathbf{Z}) - \mathbf{E}_{\mathbf{Z}} \left\{ f(\mathbf{Z}) \right\} \geq \epsilon \right\} \leq \exp \left( -\frac{2\epsilon^2}{\beta^2 \min(m, u)} \right) \quad . \qquad (3)$$

The proof of Lemma 3 appears in Appendix C.

**Remark 2** *The inequalities developed in Section 5 of Talagrand (1995) imply a concentration inequality that is similar to (3), but with worse constants.*

The inequality (2) is defined for any $(m, u)$-permutation symmetric function $f$. By specializing $f$ we obtain the following two concentration inequalities:

**Remark 3** *If $g : I_1^{m+u} \rightarrow \{0, 1\}$ and $f(\mathbf{Z}) = \frac{1}{u} \sum_{i=m+1}^{m+u} g(Z_i) - \frac{1}{m} \sum_{i=1}^{m} g(Z_i)$, then $\mathbf{E}_{\mathbf{Z}} \{ f(\mathbf{Z}) \} = 0$. Moreover, for any $i \in I_1^m$, $j \in I_{m+1}^{m+u}$, $|f(\mathbf{Z}) - f(\mathbf{Z}^{ij})| \leq \frac{1}{m} + \frac{1}{u}$. Therefore, by specializing (2) for such $f$ we obtain*

$$\mathbf{P}_{\mathbf{Z}} \left\{ \frac{1}{u} \sum_{i=m+1}^{m+u} g(Z_i) - \frac{1}{m} \sum_{i=1}^{m} g(Z_i) \geq \epsilon \right\} \leq \exp \left( -\frac{\epsilon^2 mu(m + u - 1/2)}{(m + u)^2} \cdot \frac{2\max(m, u) - 1}{\max(m, u)} \right). \qquad (4)$$

The right hand side of (4) is approximately $\exp \left( -\frac{2\epsilon^2 mu}{m+u} \right)$. The inequality (4) is an explicit (and looser) version of Vapnik's absolute bound (see El-Yaniv & Gerzon, 2005). We note that using (2) we were unable to obtain an explicit version of Vapnik's relative bound (inequality 10.14 of Vapnik, 1982).

**Remark 4** *If* $g : I_1^{m+u} \rightarrow \{0, 1\}$, $f(\mathbf{Z}) = \frac{1}{m} \sum_{i=1}^m g(Z_i)$, *then* $\mathbf{E}_{\mathbf{Z}}\{f(\mathbf{Z})\} = \frac{1}{m+u} \sum_{i=1}^{m+u} g(Z_i)$. *Moreover, for any* $i \in I_1^m$, $j \in I_{m+1}^{m+u}$, $|f(\mathbf{Z}) - f(\mathbf{Z}^{ij})| \leq \frac{1}{m}$. *Therefore, by specializing (2) for such $f$ we obtain*

$$\mathbf{P}_{\mathbf{Z}} \left\{ \frac{1}{m} \sum_{i=1}^m g(Z_i) - \frac{1}{m+u} \sum_{i=1}^{m+u} g(Z_i) \geq \epsilon \right\} \leq \exp \left( -\frac{\epsilon^2(m+u-1/2)m}{u} \cdot \frac{2\max(m,u)-1}{\max(m,u)} \right). \tag{5}$$

*The right hand side of (5) is approximately* $\exp\left(-\frac{2\epsilon^2(m+u)m}{u}\right)$. *This bound is asymptotically the same as following bound, which was developed by Serfling (1974):*

$$\mathbf{P}_{\mathbf{Z}} \left\{ \frac{1}{m} \sum_{i=1}^m g(Z_i) - \frac{1}{m+u} \sum_{i=1}^{m+u} g(Z_i) \geq \epsilon \right\} \leq \exp \left( -\frac{2\epsilon^2(m+u)m}{u+1} \right) \quad .$$

## 4. Uniform Rademacher Error Bound

In this section we develop a transductive risk bound, which is based on transductive Rademacher complexity (Definition 1). The derivation follows the standard two-step scheme, as in induction[4]:

1. Derivation of a uniform concentration inequality for a set of vectors (or functions). This inequality depends on the Rademacher complexity of the set. After substituting to the vectors (or functions) the values of the loss functions, we obtain an error bound depending on the Rademacher complexity of the values of the loss function. This step is done in Section 4.1.

2. In order to bound the Rademacher complexity in terms of the properties of the hypothesis space, the Rademacher complexity is 'translated', using its contraction property (Ledoux & Talagrand, 1991, Theorem 4.12), from the domain of loss function values to the domain of soft hypotheses from the hypothesis space. This step is done in Section 4.2.

As we show in Sections 4.1 and 4.2, the adaptation of both these steps to the transductive setting is not immediate and involves several novel ideas. In Section 4.3 we combine the results of these two steps and obtain a transductive Rademacher risk bound. We also provide a thorough comparison of our risk bound with the corresponding inductive bound.

### 4.1 Uniform Concentration Inequality for a Set of Vectors

As in induction (Koltchinskii & Panchenko, 2002), our derivation of a uniform concentration inequality for a set of vectors consists of three steps:

1. Introduction of the "ghost sample".

2. Bounding the supremum $\sup_{\mathbf{h} \in \mathcal{H}} g(\mathbf{h})$, where $g(\mathbf{h})$ is some random real-valued function, with its expectation using a concentration inequality for functions of random variables.

---

4. This scheme was introduced by Koltchinskii and Panchenko (2002). The examples of other uses of this technique can be found in the papers of Bartlett and Mendelson (2002) and Meir and Zhang (2003).

3. Bounding the expectation of the supremum using Rademacher variables.

While we follow these three steps as in induction, the establishment of each of these steps can not be achieved using inductive techniques. Throughout this section, after performing the derivation of each step in transductive context we discuss its differences from its inductive counterpart.

We introduce several new definitions. Let $\mathcal{V}$ be a set of vectors in $[B_1, B_2]^{m+u}$, $B_1 \leq 0$, $B_2 \geq 0$ and set $B \triangleq B_2 - B_1$, $B_{\max} = \max(|B_1|, |B_2|)$. Consider two independent permutations of $I_1^{m+u}$, $\mathbf{Z}$ and $\mathbf{Z}'$. For any $\mathbf{v} \in \mathcal{V}$ denote by

$$\mathbf{v}(\mathbf{Z}) \triangleq (v(Z_1), v(Z_2), \ldots, v(Z_{m+u})) \ ,$$

the vector $\mathbf{v}$ permuted according to $\mathbf{Z}$. We use the following abbreviations for averages of $\mathbf{v}$ over subsets of its components: $\mathbf{H}_k\{\mathbf{v}(\mathbf{Z})\} \triangleq \frac{1}{m} \sum_{i=1}^{k} v(Z_i)$, $\mathbf{T}_k\{\mathbf{v}(\mathbf{Z})\} \triangleq \frac{1}{u} \sum_{i=k+1}^{m+u} v(Z_i)$ (note that $\mathbf{H}$ stands for 'head' and $\mathbf{T}$, for 'tail'). In the special case where $k = m$ we set $\mathbf{H}\{\mathbf{v}(\mathbf{Z})\} \triangleq \mathbf{H}_m\{\mathbf{v}(\mathbf{Z})\}$, and $\mathbf{T}\{\mathbf{v}(\mathbf{Z})\} \triangleq \mathbf{T}_m\{\mathbf{v}(\mathbf{Z})\}$. The uniform concentration inequality that we develop shortly states that for any $\delta > 0$, with probability at least $1 - \delta$ over random permutation $\mathbf{Z}$ of $I_1^{m+u}$, for any $\mathbf{v} \in \mathcal{V}$,

$$\mathbf{T}\{\mathbf{v}(\mathbf{Z})\} \leq \mathbf{H}\{\mathbf{v}(\mathbf{Z})\} + R_{m+u}(\mathcal{V}) + O\left(\sqrt{\frac{1}{\min(m, u)} \ln \frac{1}{\delta}}\right) \ .$$

*Step 1: Introduction of the ghost sample.*

We denote by $\bar{\mathbf{v}} \triangleq \frac{1}{m+u} \sum_{i=1}^{m+u} v(i)$ the average component of $\mathbf{v}$. For any $\mathbf{v} \in \mathcal{V}$ and any permutation $\mathbf{Z}$ of $I_1^{m+u}$ we have

$$
\begin{aligned}
\mathbf{T}\{\mathbf{v}(\mathbf{Z})\} &= \mathbf{H}\{\mathbf{v}(\mathbf{Z})\} + \mathbf{T}\{\mathbf{v}(\mathbf{Z})\} - \mathbf{H}\{\mathbf{v}(\mathbf{Z})\} \\
&\leq \mathbf{H}\{\mathbf{v}(\mathbf{Z})\} + \sup_{\mathbf{v} \in \mathcal{V}} \left[ \mathbf{T}\{\mathbf{v}(\mathbf{Z})\} - \bar{\mathbf{v}} + \bar{\mathbf{v}} - \mathbf{H}\{\mathbf{v}(\mathbf{Z})\} \right] \\
&= \mathbf{H}\{\mathbf{v}(\mathbf{Z})\} + \sup_{\mathbf{v} \in \mathcal{V}} \left[ \mathbf{T}\{\mathbf{v}(\mathbf{Z})\} - \mathbf{E}_{\mathbf{Z}'} \mathbf{T}\{\mathbf{v}(\mathbf{Z}')\} + \mathbf{E}_{\mathbf{Z}'} \mathbf{H}\{\mathbf{v}(\mathbf{Z}')\} - \mathbf{H}\{\mathbf{v}(\mathbf{Z})\} \right] \\
&\leq \mathbf{H}\{\mathbf{v}(\mathbf{Z})\} + \underbrace{\mathbf{E}_{\mathbf{Z}'} \sup_{\mathbf{v} \in \mathcal{V}} \left[ \mathbf{T}\{\mathbf{v}(\mathbf{Z})\} - \mathbf{T}\{\mathbf{v}(\mathbf{Z}')\} + \mathbf{H}\{\mathbf{v}(\mathbf{Z}')\} - \mathbf{H}\{\mathbf{v}(\mathbf{Z})\} \right]}_{\triangleq \psi(\mathbf{Z})}. \quad (6)
\end{aligned}
$$

**Remark 5** *In this derivation the "ghost sample" is a permutation $\mathbf{Z}'$ of $m + u$ elements drawn from the same distribution as $\mathbf{Z}$. In inductive Rademacher-based risk bounds the ghost sample is a new training set of size $m$, independently drawn from the original one. Note that in our transductive setting the ghost sample corresponds to the independent draw of training/test set partition, which is equivalent to the independent draw of random permutation $\mathbf{Z}'$.*

**Remark 6** *In principle we could avoid the introduction of the ghost sample $\mathbf{Z}'$ and consider $m$ elements in $\mathbf{H}\{\mathbf{v}(\mathbf{Z})\}$ as ghosts of $u$ elements in $\mathbf{T}\{\mathbf{v}(\mathbf{Z})\}$. This approach would lead to* *a new definition of Rademacher averages (with $\sigma_i = -1/m$ with probability $m/(m+u)$ and $1/u$ with probability $u/(m+u)$). With this definition we can obtain Corollary 1. However, since the distribution of alternative Rademacher averages is not symmetric around zero, technically we do not know how to prove the Lemma 5 (the contraction property).*

*Step 2: Bounding the supremum with its expectation.*

Let $S \triangleq \frac{m+u}{(m+u-1/2)(1-1/(2\max(m,u)))}$. For sufficiently large $m$ and $u$, the value of $S$ is almost 1. The function $\psi(\mathbf{Z})$ is $(m, u)$-permutation symmetric in $\mathbf{Z}$. It can be verified that $|\psi(\mathbf{Z}) - \psi(\mathbf{Z}^{ij})| \leq B\left(\frac{1}{m} + \frac{1}{u}\right)$. Therefore, we can apply Lemma 2 with $\beta \triangleq B\left(\frac{1}{m} + \frac{1}{u}\right)$ to $\psi(\mathbf{Z})$. We obtain, with probability of at least $1 - \delta$ over random permutation $\mathbf{Z}$ of $I_1^{m+u}$, for all $\mathbf{v} \in \mathcal{V}$:

$$\mathbf{T}\{\mathbf{v}(\mathbf{Z})\} \leq \mathbf{H}\{\mathbf{v}(\mathbf{Z})\} + \mathbf{E}_{\mathbf{Z}}\{\psi(\mathbf{Z})\} + B\sqrt{\frac{S}{2}\left(\frac{1}{m} + \frac{1}{u}\right)\ln\frac{1}{\delta}}\ . \tag{7}$$

**Remark 7** *In induction this step is performed using an application of McDiarmid's bounded difference inequality (McDiarmid, 1989, Lemma 1.2). We cannot apply this inequality in our setting since the function under the supremum (i.e. $\psi(\mathbf{Z})$) is not a function over independent variables, but rather over permutations. Our Lemma 2 replaces the bounded difference inequality in this step.*

*Step 3: Bounding the expectation over the supremum using Rademacher random variables.*

Our goal is to bound the expectation $\mathbf{E}_{\mathbf{Z}}\{\psi(\mathbf{Z})\}$. This is done in the following lemma.

**Lemma 4** *Let $\mathbf{Z}$ be a random permutation of $I_1^{m+u}$. Let $c_0 \triangleq \sqrt{\frac{32\ln(4e)}{3}} < 5.05$. Then*

$$\mathbf{E}_{\mathbf{Z}}\{\psi(\mathbf{Z})\} \leq R_{m+u}(\mathcal{V}) + c_0 B_{\max}\left(\frac{1}{u} + \frac{1}{m}\right)\sqrt{\min(m, u)}\ .$$

**Proof:** The proof is based on ideas from the proof of Lemma 3 from Bartlett and Mendelson (2002). For technical convenience we use the following definition of pairwise Rademacher variables.

**Definition 3 (Pairwise Rademacher variables)** *Let $\mathbf{v} = (v(1), \ldots, v(m+u)) \in \mathbb{R}^{m+u}$. Let $\mathcal{V}$ be a set of vectors from $\mathbb{R}^{m+u}$. Let $\tilde{\boldsymbol{\sigma}} = \{\tilde{\sigma}_i\}_{i=1}^{m+u}$ be a vector of i.i.d. random variables defined as:*

$$\tilde{\sigma}_i = (\tilde{\sigma}_{i,1}, \tilde{\sigma}_{i,2}) = \begin{cases} \left(-\frac{1}{m}, -\frac{1}{u}\right) & \text{with probability} \quad \frac{mu}{(m+u)^2}\ ; \\ \left(-\frac{1}{m}, \frac{1}{m}\right) & \text{with probability} \quad \frac{m^2}{(m+u)^2}\ ; \\ \left(\frac{1}{u}, \frac{1}{m}\right) & \text{with probability} \quad \frac{mu}{(m+u)^2}\ ; \\ \left(\frac{1}{u}, -\frac{1}{u}\right) & \text{with probability} \quad \frac{u^2}{(m+u)^2}\ . \end{cases} \tag{8}$$

We obtain Definition 3 from Definition 1 (with $p = \frac{mu}{(m+u)^2}$) in the following way. If the Rademacher variable $\sigma_i = 1$ then we split it to $\tilde{\sigma}_i = \left(\frac{1}{u}, \frac{1}{m}\right)$. If the Rademacher variable $\sigma_i = -1$ then we split it to $\tilde{\sigma}_i = \left(-\frac{1}{m}, -\frac{1}{u}\right)$. If the Rademacher variable $\sigma_i = 0$ then we split it randomly to $\left(-\frac{1}{m}, \frac{1}{m}\right)$ or $\left(\frac{1}{u}, -\frac{1}{u}\right)$. The first component of $\tilde{\sigma}_i$ indicates if the $i$th component of $\mathbf{v}$ is in the first elements of $\mathbf{v}(\mathbf{Z})$ or in the last $u$ elements of $\mathbf{v}(\mathbf{Z})$. If the former case the value of $\tilde{\sigma}_i$ is $-\frac{1}{m}$ and in the latter case the value of $\tilde{\sigma}_i$ is $\frac{1}{u}$. The second component of $\tilde{\sigma}_i$ has the same meaning as the first one, but with $\mathbf{Z}$ replaced by $\mathbf{Z}'$.

The values $\pm\frac{1}{m}$ and $\pm\frac{1}{u}$ are exactly the coefficients appearing inside $\mathbf{T}\{\mathbf{v}(\mathbf{Z})\}$, $\mathbf{T}\{\mathbf{v}(\mathbf{Z}')\}$, $\mathbf{H}\{\mathbf{v}(\mathbf{Z}')\}$ and $\mathbf{H}\{\mathbf{v}(\mathbf{Z})\}$ in (6). These coefficients are random and their distribution is induced by the uniform distribution over permutations. In the course of the proof we will establish the precise relation between the distribution of $\pm\frac{1}{m}$ and $\pm\frac{1}{u}$ coefficients and the distribution (8) of pairwise Rademacher variables.

It is easy to verify that

$$R_{m+u}(\mathcal{V}) = \mathbf{E}_{\tilde{\boldsymbol{\sigma}}} \left\{ \sup_{\mathbf{v} \in \mathcal{V}} \sum_{i=1}^{m+u} (\tilde{\sigma}_{i,1} + \tilde{\sigma}_{i,2}) v(i) \right\}. \tag{9}$$

Let $n_1, n_2$ and $n_3$ be the number of random variables $\tilde{\sigma}_i$ realizing the value $\left(-\frac{1}{m}, -\frac{1}{u}\right)$, $\left(-\frac{1}{m}, \frac{1}{m}\right)$, $\left(\frac{1}{u}, \frac{1}{m}\right)$, respectively. Set $N_1 \triangleq n_1 + n_2$ and $N_2 \triangleq n_2 + n_3$. Note that the $n_i$'s and $N_i$'s are random variables. Denote by $\mathtt{Rad}$ the distribution of $\tilde{\boldsymbol{\sigma}}$ defined by (8) and by $\mathtt{Rad}(N_1, N_2)$, the distribution $\mathtt{Rad}$ conditioned on the events $n_1 + n_2 = N_1$ and $n_2 + n_3 = N_2$. We define

$$s(N_1, N_2) \triangleq \mathbf{E}_{\tilde{\boldsymbol{\sigma}} \sim \mathtt{Rad}(N_1, N_2)} \left\{ \sup_{\mathbf{v} \in \mathcal{V}} \sum_{i=1}^{m+u} (\tilde{\sigma}_{i,1} + \tilde{\sigma}_{i,2}) v(i) \right\}.$$

The rest of the proof is based on the following three claims:

**Claim 1.** $R_{m+u}(\mathcal{V}) = \mathbf{E}_{N_1, N_2}\{s(N_1, N_2)\}$.

**Claim 2.** $\mathbf{E}_{\mathbf{Z}}\{\psi(\mathbf{Z})\} = s(\mathbf{E}_{\tilde{\boldsymbol{\sigma}}} N_1, \mathbf{E}_{\tilde{\boldsymbol{\sigma}}} N_2)$.

**Claim 3.** $s(\mathbf{E}_{\tilde{\boldsymbol{\sigma}}} N_1, \mathbf{E}_{\tilde{\boldsymbol{\sigma}}} N_2) - \mathbf{E}_{N_1, N_2}\{s(N_1, N_2)\} \le c_0 B_{\max} \left(\frac{1}{u} + \frac{1}{m}\right) \sqrt{m}$.

Having established these three claims we immediately obtain

$$\mathbf{E}_{\mathbf{Z}}\{g(\mathbf{Z})\} \le \tilde{R}_{m+u}(\mathcal{V}) + c_0 B_{\max} \left(\frac{1}{u} + \frac{1}{m}\right) \sqrt{m} . \tag{10}$$

The entire development is symmetric in $m$ and $u$ and, therefore, we also obtain the same result but with $\sqrt{u}$ instead of $\sqrt{m}$. By taking the minimum of (10) and the symmetric bound (with $\sqrt{u}$) we establish the theorem.

The proof of the above three claims appears in Appendix D. $\qquad\square$

**Remark 8** *The technique we use to bound the expectation of the supremum is more complicated than the technique of Koltchinskii and Panchenko (2002) that is commonly used in induction. This is caused by the structure of the function under the supremum (i.e., $g(\mathbf{Z})$). From a conceptual point of view, this step utilizes our novel definition of transductive Rademacher complexity.*

By combining (7) and Lemma 4 we obtain the next concentration inequality, which is the main result of this section.

**Theorem 1** *Let $B_1 \leq 0$, $B_2 \geq 0$ and $\mathcal{V}$ be a (possibly infinite) set of real-valued vectors in $[B_1, B_2]^{m+u}$. Let $B \triangleq B_2 - B_1$ and $B_{\max} \triangleq \max(|B_1|, |B_2|)$. Let $Q \triangleq \left(\frac{1}{u} + \frac{1}{m}\right)$, $S \triangleq \frac{m+u}{(m+u-1/2)(1-1/2(\max(m,u)))}$ and $c_0 \triangleq \sqrt{\frac{32\ln(4e)}{3}} < 5.05$. Then with probability of at least $1 - \delta$ over random permutation $\mathbf{Z}$ of $I_1^{m+u}$, for all $\mathbf{v} \in \mathcal{V}$,*

$$\mathbf{T}\{\mathbf{v}(\mathbf{Z})\} \leq \mathbf{H}\{\mathbf{v}(\mathbf{Z})\} + R_{m+u}(\mathcal{V}) + B_{\max}c_0 Q\sqrt{\min(m,u)} + B\sqrt{\frac{S}{2}Q\ln\frac{1}{\delta}}. \quad (11)$$

We defer the analysis of the slack terms $B_{\max}c_0 Q\sqrt{\min(m,u)}$ and $B\sqrt{\frac{S}{2}Q\ln\frac{1}{\delta}}$ to Section 4.3. We now instantiate the inequality (11) to obtain our first risk bound. The idea is to apply Theorem 1 with an appropriate instantiation of the set $\mathcal{V}$ so that $\mathbf{T}\{\mathbf{v}(\mathbf{Z})\}$ will correspond to the test error and $\mathbf{H}\{\mathbf{v}(\mathbf{Z})\}$ to the empirical error. For a true (unknown) labeling of the full-sample $\mathbf{y}$ and any $\mathbf{h} \in \mathcal{H}_{\text{out}}$ we define

$$\boldsymbol{\ell}^{\mathbf{y}}(\mathbf{h}) \triangleq (\ell(h(1), y_1), \ldots, \ell(h(m+u), y_{m+u}))$$

and set $L_\mathcal{H} = \{\mathbf{v} \; : \; \mathbf{v} = \boldsymbol{\ell}^{\mathbf{y}}(\mathbf{h}), \; \mathbf{h} \in \mathcal{H}_{\text{out}}\}$. Thus $\boldsymbol{\ell}^{\mathbf{y}}(\mathbf{h})$ is a vector of the values of the 0/1 loss over all full sample examples, when transductive algorithm is operated on some training/test partition. The set $L_\mathcal{H}$ is the set of all possible vectors $\boldsymbol{\ell}^{\mathbf{y}}(\mathbf{h})$, over all possible training/test partitions. We apply Theorem 1 with $\mathcal{V} \triangleq L_\mathcal{H}$, $\mathbf{v} \triangleq \boldsymbol{\ell}(\mathbf{h})$, $B_{\max} = B = 1$ and obtain the following corollary:

**Corollary 1** *Let $Q$, $S$ and $c_0$ be as defined in Theorem 1. For any $\delta > 0$, with probability of at least $1 - \delta$ over the choice of the training set from $X_{m+u}$, for all $\mathbf{h} \in \mathcal{H}_{out}$,*

$$\mathcal{L}_u(\mathbf{h}) \leq \widehat{\mathcal{L}}_m(\mathbf{h}) + R_{m+u}(L_\mathcal{H}) + B_{\max}c_0 Q\sqrt{\min(m,u)} + \sqrt{\frac{S}{2}Q\ln\frac{1}{\delta}} \; . \quad (12)$$

We defer the analysis of the slack terms $B_{\max}c_0 Q\sqrt{\min(m,u)}$ and $B\sqrt{\frac{S}{2}Q\ln\frac{1}{\delta}}$ to Section 4.3. While the bound (12) is obtained by a straightforward application of the concentration inequality (11), it is not convenient to deal with. That's because it is not clear how to bound the Rademacher complexity $R_{m+u}(L_\mathcal{H})$ of the 0/1 loss values in terms of the properties of transductive algorithm. In the next sections we eliminate this deficiency by utilizing margin loss function.

## 4.2 Contraction of Rademacher Complexity

The following lemma is a version of the well-known 'contraction principle' of the theory of Rademacher averages (see Theorem 4.12 of Ledoux & Talagrand, 1991, and Ambroladze, Parrado-Hernandez, & Shawe-Taylor, 2007). The lemma is an adaptation, which accommodates the transductive Rademacher variables, of Lemma 5 of Meir and Zhang (2003). The proof is provided in Appendix E.

**Lemma 5** *Let $\mathcal{V} \subseteq \mathbb{R}^{m+u}$ be a set of vectors. Let $f$ and $g$ be real-valued functions. Let $\boldsymbol{\sigma} = \{\sigma_i\}_{i=1}^{m+u}$ be Rademacher variables, as defined in (1). If for all $1 \leq i \leq m+u$ and any*

$\mathbf{v}, \mathbf{v}' \in \mathcal{V}$, $|f(v_i) - f(v'_i)| \leq |g(v_i) - g(v'_i)|$, then

$$\mathbf{E}_{\boldsymbol{\sigma}} \sup_{\mathbf{v} \in \mathcal{V}} \left[ \sum_{i=1}^{m+u} \sigma_i f(v_i) \right] \leq \mathbf{E}_{\boldsymbol{\sigma}} \sup_{\mathbf{v} \in \mathcal{V}} \left[ \sum_{i=1}^{m+u} \sigma_i g(v_i) \right] .$$

Let $\mathbf{y} = (y_1, \ldots, y_{m+u}) \in \mathbb{R}^{m+u}$ be a true (unknown) labeling of the full-sample. Similarly to what was done in the derivation of Corollary 1, for any $\mathbf{h} \in \mathcal{H}_{\text{out}}$ we define $\ell_\gamma^\mathcal{Y}(h(i)) \triangleq \ell_\gamma(h(i), y_i)$ and

$$\boldsymbol{\ell}_\gamma^\mathcal{Y}(\mathbf{h}) \triangleq (\ell_\gamma^\mathcal{Y}(h(1)), \ldots, \ell_\gamma^\mathcal{Y}(h(m+u)))$$

and set $L_\mathcal{H}^\gamma = \{\mathbf{v} : \mathbf{v} = \boldsymbol{\ell}_\gamma^\mathcal{Y}(\mathbf{h}), \mathbf{h} \in \mathcal{H}_{\text{out}}\}$. Noting that $\ell_\gamma^\mathcal{Y}$ satisfies the Lipschitz condition $|\ell_\gamma^\mathcal{Y}(h(i)) - \ell_\gamma^\mathcal{Y}(h'(i))| \leq \frac{1}{\gamma} |h(i) - h'(i)|$, we apply Lemma 5 with $\mathcal{V} \triangleq L_\mathcal{H}^\gamma$, $f(v_i) \triangleq \ell_\gamma^\mathcal{Y}(h(i))$ and $g(v_i) \triangleq h(i)/\gamma$, to get

$$\mathbf{E}_{\boldsymbol{\sigma}} \left\{ \sup_{\mathbf{h} \in \mathcal{H}_{\text{out}}} \sum_{i=1}^{m+u} \sigma_i \ell_\gamma^\mathcal{Y}(h(i)) \right\} \leq \frac{1}{\gamma} \mathbf{E}_{\boldsymbol{\sigma}} \left\{ \sup_{\mathbf{h} \in \mathcal{H}_{\text{out}}} \sum_{i=1}^{m+u} \sigma_i h(i) \right\} . \tag{13}$$

It follows from (13) that

$$R_{m+u}(L_\mathcal{H}^\gamma) \leq \frac{1}{\gamma} R_{m+u}(\mathcal{H}_{\text{out}}) . \tag{14}$$

### 4.3 Risk Bound and Comparison with Related Results

Applying Theorem 1 with $\mathcal{V} \triangleq L_\mathcal{H}^\gamma$, $\mathbf{v} \triangleq \boldsymbol{\ell}_\gamma(\mathbf{h})$, $B_{\max} = B = 1$, and using the inequality (14) we obtain[5]:

**Theorem 2** *Let $\mathcal{H}_{\text{out}}$ be the set of full-sample soft labelings of the algorithm, generated by operating it on all possible training/test set partitions. The choice of $\mathcal{H}_{\text{out}}$ can depend on the full-sample $X_{m+u}$. Let $c_0 = \sqrt{\frac{32 \ln(4e)}{3}} < 5.05$, $Q \triangleq \left( \frac{1}{u} + \frac{1}{m} \right)$ and $S \triangleq \frac{m+u}{(m+u-1/2)(1-1/(2\max(m,u)))}$. For any fixed $\gamma$, with probability of at least $1 - \delta$ over the choice of the training set from $X_{m+u}$, for all $\mathbf{h} \in \mathcal{H}_{\text{out}}$,*

$$\mathcal{L}_u(\mathbf{h}) \leq \mathcal{L}_u^\gamma(\mathbf{h}) \leq \widehat{\mathcal{L}}_m^\gamma(\mathbf{h}) + \frac{R_{m+u}(\mathcal{H}_{\text{out}})}{\gamma} + c_0 Q \sqrt{\min(m, u)} + \sqrt{\frac{SQ}{2} \ln \frac{1}{\delta}} . \tag{15}$$

For large enough values of $m$ and $u$ the value of $S$ is close to 1. Therefore the slack term $c_0 Q \sqrt{\min(m, u)} + \sqrt{\frac{S}{2} Q \ln \frac{1}{\delta}}$ is of order $O\left( 1/\sqrt{\min(m, u)} \right)$. The convergence rate of $O\left( 1/\sqrt{\min(m, u)} \right)$ can be very slow if $m$ is very small or $u \ll m$. Slow rate for small $m$ is not surprising, but a latter case of $u \ll m$ is somewhat surprising. However note that if $u \ll m$ then the mean $\mu$ of $u$ elements, drawn from $m + u$ elements, has a large variance. Hence, in this case any high-confidence interval for the estimation of $\mu$ will be large. This confidence interval is reflected in the slack term of (15).

---

5. This bound holds for any *fixed* margin parameter $\gamma$. Using the technique of the proof of Theorem 18 of Bousquet and Elisseeff (2002), we can also obtain a bound that is uniform in $\gamma$.

We now compare the bound (15) with the Rademacher-based inductive risk bounds. We use the following variant of Rademacher-based inductive risk bound of Meir and Zhang (2003):

**Theorem 3** *Let $\mathcal{D}$ be a probability distribution over $\mathcal{X}$. Suppose that a set of examples $S_m = \{(x_i, y_i)\}_{i=1}^m$ is sampled i.i.d. from $\mathcal{X}$ according to $\mathcal{D}$. Let $\mathcal{F}$ be a class of functions each maps $\mathcal{X}$ to $\mathbb{R}$ and $\widehat{R}_m^{(\text{ind})}(\mathcal{F})$ be the empirical Rademacher complexity of $\mathcal{F}$ (Definition 2). Let $\mathcal{L}(f) = \mathbf{E}_{(x,y)\sim\mathcal{D}}\{\ell(f(x),y)\}$ and $\widehat{\mathcal{L}}^\gamma(f) = \frac{1}{m}\sum_{i=1}^m \ell_\gamma(f(x_i),y_i)$ be respectively the 0/1 generalization error and empirical margin error of $f$. Then for any $\delta > 0$ and $\gamma > 0$, with probability of at least $1 - \delta$ over the random draw of $S_m$, for any $f \in \mathcal{F}$,*

$$\mathcal{L}(f) \leq \widehat{\mathcal{L}}^\gamma(f) + \frac{\widehat{R}_m^{(\text{ind})}(\mathcal{F})}{\gamma} + \sqrt{\frac{2\ln(2/\delta)}{m}} \quad . \tag{16}$$

The slack term in the bound (16) is of order $O(1/\sqrt{m})$. The bounds (15) and (16) are not quantitatively comparable. The inductive bound holds with high probability over the random selection of $m$ examples from some distribution $\mathcal{D}$. This bound is on average (generalization) error, over all examples in $\mathcal{D}$. The transductive bound holds with high probability over the random selection of a training/test partition. This bound is on the test error of some hypothesis over a particular set of $u$ points.

A kind of meaningful comparison can be obtained as follows. Using the given full (transductive) sample $X_{m+u}$, we define a corresponding inductive distribution $\mathcal{D}_{\text{trans}}$ as the uniform distribution over $X_{m+u}$; that is, a training set of size $m$ will be generated by sampling from $X_{m+u}$ $m$ times with replacements. Given an inductive hypothesis space $\mathcal{F} = \{f\}$ of function we define the transductive hypothesis space $\mathcal{H}_{\mathcal{F}}$ as a projection of $\mathcal{F}$ into the full sample $X_{m+u}$: $\mathcal{H}_{\mathcal{F}} = \{\mathbf{h} \in \mathbb{R}^{m+u} \ : \ \exists f \in \mathcal{F}, \forall 1 \leq i \leq m+u, h(i) = f(x_i)\}$. By such definition of $\mathcal{H}_{\mathcal{F}}$, $\mathcal{L}(f) = \mathcal{L}_{m+u}(\mathbf{h})$.

Our final step towards a meaningful comparison would be to translate a transductive bound of the form $\mathcal{L}_u(\mathbf{h}) \leq \widehat{\mathcal{L}}_m^\gamma(\mathbf{h}) + \text{slack}$ to a bound on the average error of the hypothesis[6] $\mathbf{h}$:

$$\begin{aligned}
\mathcal{L}_{m+u}(\mathbf{h}) &\leq \mathcal{L}_{m+u}^\gamma(\mathbf{h}) = \frac{m\widehat{\mathcal{L}}_m^\gamma(\mathbf{h}) + u\mathcal{L}_u^\gamma(\mathbf{h})}{m+u} \leq \frac{m\widehat{\mathcal{L}}_m^\gamma(\mathbf{h}) + u\left(\widehat{\mathcal{L}}_m^\gamma(\mathbf{h}) + \text{slack}\right)}{m+u} \\
&= \widehat{\mathcal{L}}_m^\gamma(\mathbf{h}) + \frac{u}{m+u} \cdot \text{slack}
\end{aligned} \tag{17}$$

We instantiate (17) to the bound (15) and obtain

$$\mathcal{L}_{m+u}(\mathbf{h}) \leq \widehat{\mathcal{L}}_m^\gamma(\mathbf{h}) + \frac{u}{m+u}\frac{R_{m+u}(\mathcal{H}_{\mathcal{F}})}{\gamma} + \frac{u}{m+u}\left[c_0 Q\sqrt{\min(m,u)} + \sqrt{\frac{SQ}{2}\ln\frac{1}{\delta}}\right] . \tag{18}$$

---

6. Alternatively, to compare (15) and (16), we could try to express the bound (16) as the bound on the error of $f$ on $X_u$ (the randomly drawn subset of $u$ examples). The bound (16) holds for the setting of random draws with replacement. In this setting the number of unique training examples can be smaller than $m$ and thus the number of the remaining test examples is larger than $u$. Hence the draw of $m$ training examples with replacement does not imply the draw of the subset of $u$ test examples, as in transductive setting. Thus we cannot express the bound (16) as the bound on the randomly drawn $X_u$

Now given a transductive problem we consider the corresponding inductive bound obtained from (16) under the distribution $\mathcal{D}_{\text{trans}}$ and compare it to the bound (18).

Note that in the inductive bound (16) the sampling of the training set is done with replacement, while in the transductive bound (18) it is done without replacement. Thus, in the inductive case the actual number of distinct training examples may be smaller than $m$.

The bounds (16) and (18) consist of three terms: empirical error term (first summand in (16) and (18)), the term depending on the Rademacher complexity (second summand in (16) and (18)) and the slack term (third summand in (16) and third and fourth summands in (18)). The empirical error terms are the same in both bounds. It is hard to compare analytically the Rademacher complexity terms. This is because the inductive bound is derived for the setting of sampling with replacement and the transductive bound is derived for the setting of sampling without replacement. Thus, in the transductive Rademacher complexity each example $x_i \in X_{m+u}$ appears in $R_{m+u}(\mathcal{H}_{\text{out}})$ only once and is multiplied by $\sigma_i$. In contrast, due to the sampling with replacement, in the inductive Rademacher term the example $x_i \in X_{m+u}$ can appear several times in $\widehat{R}_{m+u}^{(\text{ind})}(\mathcal{F})$, multiplied by different values of the Rademacher variables.

Nevertheless, in transduction we have a full control over the Rademacher complexity (since we can choose $\mathcal{H}_{\text{out}}$ after observing the full sample $X_{m+u}$) and can choose a hypothesis space $\mathcal{H}_{\text{out}}$ with arbitrarily small Rademacher complexity. In induction we choose $\mathcal{F}$ before observing any data. Hence, if we are lucky with the full sample $X_{m+u}$ then $\widehat{R}_{m+u}^{(\text{ind})}(\mathcal{F})$ is small, and if we are unlucky with $X_{m+u}$ then $\widehat{R}_{m+u}^{(\text{ind})}(\mathcal{F})$ can be large. Thus, under these provisions we can argue that the transductive Rademacher term is not larger than the inductive counterpart.

Finally, we compare the slack terms in (16) and (18). If $m \approx u$ or $m \ll u$ then the slack term of (18) is of order $O\left(1/\sqrt{m}\right)$, which is the same as the corresponding term in (16). But if $m \gg u$ then the slack term of (18) is of order $O\left(1/(m\sqrt{u})\right)$, which is much smaller than $O(1/\sqrt{m})$ of the slack term in (16).

Based on the comparison of the corresponding terms in (16) and (18) our conclusion is that in the regime of $u \ll m$ the transductive bound is significantly tighter than the inductive one.[7]

## 5. Unlabeled-Labeled Representation (ULR) of Transductive Algorithms

Let $r$ be any natural number and let $U$ be an $(m+u) \times r$ matrix depending only on $X_{m+u}$. Let $\boldsymbol{\alpha}$ be an $r \times 1$ vector that may depend on both $S_m$ and $X_u$. The soft classification output $\mathbf{h}$ of any transductive algorithm can be represented by

$$\mathbf{h} = U \cdot \boldsymbol{\alpha} \ . \tag{19}$$

We refer to (19) as an *unlabeled-labeled representation (ULR)*. In this section we develop bounds on the Rademacher complexity of algorithms based on their ULRs. We note that any transductive algorithm has a trivial ULR, for example, by taking $r = m+u$, setting $U$ to be the identity matrix and assigning $\boldsymbol{\alpha}$ to any desired (soft) labels. We are interested in "non-trivial" ULRs and provide useful bounds for such representations.[8]

---

7. The regime of $u \ll m$ occurs in the following class of applications. Given a large library of tagged objects, the goal of the learner is to assign the tags to a small quantity of the newly arrived objects. The example of such application is the organization of daily news.

In a "vanilla" ULR, $U$ is an $(m+u) \times (m+u)$ matrix and $\boldsymbol{\alpha} = (\alpha_1, \ldots, \alpha_{m+u})$ simply specifies the given labels in $S_m$ (where $\alpha_i = y_i$ for labeled points, and $\alpha_i = 0$ otherwise). From our point of view any vanilla ULR is not trivial because $\boldsymbol{\alpha}$ does not encode the final classification of the algorithm. For example, the algorithm of Zhou et al. (2004) straightforwardly admits a vanilla ULR. On the other hand, the natural (non-trivial) ULR of the algorithms of Zhu et al. (2003) and Belkin and Niyogi (2004) are not of the vanilla type. For some algorithms it is not necessarily obvious how to find non-trivial ULRs. In Sections 6 we consider two such cases – in particular, the algorithms of Joachims (2003) and Belkin et al. (2004).

The rest of this section is organized as follows. In Section 5.1 we present a generic bound on the Rademacher complexity of any transductive algorithm based on its ULR. In Section 5.2 we consider a case when the matrix $U$ is a kernel matrix. For this case we develop another bound on the transductive Rademacher complexity. Finally, in Section 5.3 we present a method of computing high-confidence estimate of the transductive Rademacher complexity.

## 5.1 Generic Bound on Transductive Rademacher Complexity

We now present a bound on the transductive Rademacher complexity of any transductive algorithm based on its ULR. Let $\{\lambda_i\}_{i=1}^r$ be the singular values of $U$. We use the well-known fact that $\|U\|_{\mathrm{Fro}} = \sqrt{\sum_{i=1}^r \lambda_i^2}$, where $\|U\|_{\mathrm{Fro}} \triangleq \sqrt{\sum_{i,j}(U(i,j))^2}$ is the Frobenius norm of $U$. Suppose that $\|\boldsymbol{\alpha}\|_2 \leq \mu_1$ for some $\mu_1$. Let $\mathcal{H}_{\mathrm{out}} \triangleq \mathcal{H}_{\mathrm{out}}(U)$ be the set of all possible outputs of the algorithm when operated on all possible training/test set partitions of the full-sample $X_{m+u}$. Let $Q \triangleq \frac{1}{m} + \frac{1}{u}$. Using the abbreviation $U(i, \cdot)$ for the $i$th row of $U$ and following the proof idea of Lemma 22 of Bartlett and Mendelson (2002), we have that

$$
\begin{aligned}
R_{m+u}(\mathcal{H}_{\mathrm{out}}) &= Q \cdot \mathbf{E}_{\boldsymbol{\sigma}} \left\{ \sup_{\mathbf{h} \in \mathcal{H}_{\mathrm{out}}} \sum_{i=1}^{m+u} \sigma_i h(x_i) \right\} = Q \cdot \mathbf{E}_{\boldsymbol{\sigma}} \left\{ \sup_{\boldsymbol{\alpha}: \|\boldsymbol{\alpha}\|_2 \leq \mu_1} \sum_{i=1}^{m+u} \sigma_i \langle \boldsymbol{\alpha}, U(i, \cdot) \rangle \right\} \\
&= Q \cdot \mathbf{E}_{\boldsymbol{\sigma}} \left\{ \sup_{\boldsymbol{\alpha}: \|\boldsymbol{\alpha}\|_2 \leq \mu_1} \langle \boldsymbol{\alpha}, \sum_{i=1}^{m+u} \sigma_i U(i, \cdot) \rangle \right\} \\
&= Q \mu_1 \mathbf{E}_{\boldsymbol{\sigma}} \left\{ \left\| \sum_{i=1}^{m+u} \sigma_i U(i, \cdot) \right\|_2 \right\} \\
&= Q \mu_1 \mathbf{E}_{\boldsymbol{\sigma}} \left\{ \sqrt{\sum_{i,j=1}^{m+u} \sigma_i \sigma_j \langle U(i, \cdot), U(j, \cdot) \rangle} \right\} \\
&\leq Q \mu_1 \sqrt{\sum_{i,j=1}^{m+u} \mathbf{E}_{\boldsymbol{\sigma}} \left\{ \sigma_i \sigma_j \langle U(i, \cdot), U(j, \cdot) \rangle \right\}}
\end{aligned}
$$

(20)

(21)

---

8. For the trivial representation where $U$ is the identity matrix multiplied by constant we show in Lemma 6 that the risk bound (15), combined with the forthcoming Rademacher complexity bound (22), is greater than 1.

$$= \mu_1 \sqrt{\sum_{i=1}^{m+u} \frac{2}{mu} \langle U(i,\cdot), U(i,\cdot) \rangle} = \mu_1 \sqrt{\frac{2}{mu} \|U\|_{\mathrm{Fro}}^2} = \mu_1 \sqrt{\frac{2}{mu} \sum_{i=1}^{r} \lambda_i^2} \ . \qquad (22)$$

where (20) and (21) are obtained using, respectively, the Cauchy-Schwarz and Jensen inequalities. Using the bound (22) in conjunction with Theorem 2 we immediately get a data-dependent error bound for any algorithm, which can be computed once we derive an upper bound on the maximal length of possible values of the $\boldsymbol{\alpha}$ vector, appearing in its ULR. Notice that for any vanilla ULR (and thus for the "consistency method" of Zhou et al. (2004)), $\mu_1 = \sqrt{m}$. In Section 6 we derive a tight bound on $\mu_1$ for non-trivial ULRs of SGT of Joachims (2003) and of the consistency method of Zhou et al. (2004).

The bound (22) is syntactically similar in form to a corresponding inductive Rademacher bound for kernel machines (Bartlett & Mendelson, 2002). However, as noted above, the fundamental difference is that in induction, the choice of the kernel (and therefore $\mathcal{H}_{\mathrm{out}}$) must be *data-independent* in the sense that it must be selected *before* the training examples are observed. In our transductive setting, $U$ and $\mathcal{H}_{\mathrm{out}}$ can be selected *after* the unlabeled full-sample is observed.

The Rademacher bound (22), as well as the forthcoming Rademacher bound (25), depend on the spectrum of the matrix $U$. As we will see in Section 6, in non-trivial ULRs of some transductive algorithms (the algorithms of Zhou et al., 2004 and of Belkin et al., 2004) the spectrum of $U$ depends on the spectrum of the Laplacian of the graph used by the algorithm. Thus by transforming the spectrum of Laplacian we control the Rademacher complexity of the hypothesis class. There exists strong empirical evidence (see Chapelle et al., 2003; Joachims, 2003; Johnson & Zhang, 2008) that such spectral transformations improve the performance of the transductive algorithms.

The next lemma (proven in Appendix F) shows that for "trivial" ULRs the resulting risk bound is vacuous.

**Lemma 6** *Let $\boldsymbol{\alpha} \in \mathbb{R}^{m+u}$ be a vector depending on both $S_m$ and $X_u$. Let $c \in \mathbb{R}$, $U \overset{\triangle}{=} c \cdot I$ and $\mathcal{A}$ be transductive algorithm generating soft-classification vector $\mathbf{h} = U \cdot \boldsymbol{\alpha}$. Let $\{\lambda_i\}_{i=1}^{r}$ be the singular values of $U$ and $\mu_1$ be the upper bound on $\|\boldsymbol{\alpha}\|_2$. For the algorithm $\mathcal{A}$ the bound (22) in conjunction with the bound (15) is vacuous; namely, for any $\gamma \in (0,1)$ and any $\mathbf{h}$ generated by $\mathcal{A}$ it holds that*

$$\widehat{\mathcal{L}}_m^{\gamma}(\mathbf{h}) + \frac{\mu_1}{\gamma} \sqrt{\frac{2}{mu} \sum_{i=1}^{k} \lambda_i^2} + c_0 Q \sqrt{\min(m,u)} + \sqrt{\frac{S}{2} Q \ln \frac{1}{\delta}} \geq 1 \ .$$

## 5.2 Kernel ULR

If $r = m + u$ and the matrix $U$ is a kernel matrix (this holds if $U$ is positive semidefinite), then we say that the decomposition is a *kernel-ULR*. Let $\mathcal{G} \subseteq \mathbb{R}^{m+u}$ be the reproducing kernel Hilbert space (RKHS), corresponding to $U$. We denote by $\langle \cdot, \cdot \rangle_{\mathcal{G}}$ the inner product in $\mathcal{G}$. Since $U$ is a kernel matrix, by the reproducing property[9] of $\mathcal{G}$, $U(i,j) = \langle U(i,\cdot), U(j,\cdot) \rangle_{\mathcal{G}}$.

---

9. This means that for all $\mathbf{h} \in \mathcal{G}$ and $i \in I_1^{m+u}$, $h(i) = \langle U(i,\cdot), \mathbf{h} \rangle_{\mathcal{G}}$.

Suppose that the vector $\boldsymbol{\alpha}$ satisfies $\sqrt{\boldsymbol{\alpha}^T U \boldsymbol{\alpha}} \leq \mu_2$ for some $\mu_2$. Let $\{\lambda_i\}_{i=1}^{m+u}$ be the eigenvalues of $U$. By similar arguments used to derive (22) we have:

$$
\begin{aligned}
R_{m+u}(\mathcal{H}_{\text{out}}) &= Q \cdot \mathbf{E}_{\boldsymbol{\sigma}}\left\{\sup_{\mathbf{h}\in\mathcal{H}_{\text{out}}}\sum_{i=1}^{m+u}\sigma_i h(x_i)\right\} = Q \cdot \mathbf{E}_{\boldsymbol{\sigma}}\left\{\sup_{\boldsymbol{\alpha}}\sum_{i=1}^{m+u}\sigma_i\sum_{j=1}^{m+u}\alpha_j U(i,j)\right\} \\
&= Q \cdot \mathbf{E}_{\boldsymbol{\sigma}}\left\{\sup_{\boldsymbol{\alpha}}\sum_{i=1}^{m+u}\sigma_i\sum_{j=1}^{m+u}\alpha_j\langle U(i,\cdot), U(j,\cdot)\rangle_{\mathcal{G}}\right\} \\
&= Q \cdot \mathbf{E}_{\boldsymbol{\sigma}}\left\{\sup_{\boldsymbol{\alpha}}\left\langle\sum_{i=1}^{m+u}\sigma_i U(i,\cdot), \sum_{j=1}^{m+u}\alpha_j U(j,\cdot)\right\rangle_{\mathcal{G}}\right\} \\
&\leq Q \cdot \mathbf{E}_{\boldsymbol{\sigma}}\left\{\sup_{\boldsymbol{\alpha}}\left\|\sum_{i=1}^{m+u}\sigma_i U(i,\cdot)\right\|_{\mathcal{G}} \cdot \left\|\sum_{j=1}^{m+u}\alpha_j U(j,\cdot)\right\|_{\mathcal{G}}\right\} \quad (23)\\
&= Q\mu_2\mathbf{E}_{\boldsymbol{\sigma}}\left\{\left\|\sum_{i=1}^{m+u}\sigma_i U(i,\cdot)\right\|_{\mathcal{G}}\right\} \\
&= Q\mu_2\mathbf{E}_{\boldsymbol{\sigma}}\left\{\sqrt{\left\langle\sum_{i=1}^{m+u}\sigma_i U(i,\cdot), \sum_{j=1}^{m+u}\sigma_j U(j,\cdot)\right\rangle_{\mathcal{G}}}\right\} \\
&= Q\mu_2\mathbf{E}_{\boldsymbol{\sigma}}\left\{\sqrt{\sum_{i,j=1}^{m+u}\sigma_i\sigma_j U(i,j)}\right\} \leq Q\mu_2\sqrt{\sum_{i,j=1}^{m+u}\mathbf{E}_{\boldsymbol{\sigma}}\left\{\sigma_i\sigma_j U(i,j)\right\}} \quad (24)\\
&= \mu_2\sqrt{\sum_{i=1}^{m+u}\frac{2}{mu}U(i,i)} = \mu_2\sqrt{\frac{2\cdot\text{trace}(U)}{mu}} = \mu_2\sqrt{\frac{2}{mu}\sum_{i=1}^{m+u}\lambda_i} \ . \quad (25)
\end{aligned}
$$

The inequalities (23) and (24) are obtained using, respectively, Cauchy-Schwarz and Jensen inequalities. Finally, the first equality in (25) follows from the definition of Rademacher variables (see Definition 1).

If transductive algorithm has kernel-ULR then we can use both (25) and (22) to bound its Rademacher complexity. The kernel bound (25) can be tighter than its non-kernel counterpart (22) when the kernel matrix has eigenvalues larger than one and/or $\mu_2 < \mu_1$. In Section 6 we derive a tight bound on $\mu_1$ for non-trivial ULRs of "consistency method" of Zhou et al. (2004) and of the Tikhonov regularization method of Belkin et al. (2004).

## 5.3 Monte-Carlo Rademacher Bounds

We now show how to compute Monte-Carlo Rademacher bounds with high confidence for any transductive algorithm using its ULR. Our empirical examination of these bounds (see Section 6.3) shows that they are tighter than the analytical bounds (22) and (25). The technique, which is based on a simple application of Hoeffding's inequality, is made particularly simple for vanilla ULRs.

Let $\mathcal{V} \subseteq \mathbb{R}^{m+u}$ be a set of vectors, $Q \triangleq \frac{1}{m} + \frac{1}{u}$, $\boldsymbol{\sigma} \in \mathbb{R}^{m+u}$ to be a Rademacher vector (1), and $g(\boldsymbol{\sigma}) = \sup_{\mathbf{v} \in \mathcal{V}} \boldsymbol{\sigma}^T \cdot \mathbf{v}$. By Definition 1, $R_{m+u}(\mathcal{V}) = Q \cdot \mathbf{E}_{\boldsymbol{\sigma}}\{g(\boldsymbol{\sigma})\}$. Let $\boldsymbol{\sigma_1}, \ldots, \boldsymbol{\sigma_n}$ be an i.i.d. sample of Rademacher vectors. We estimate $R_{m+u}(\mathcal{V})$ with high confidence by applying the Hoeffding inequality on $\sum_{i=1}^{n} \frac{1}{n} g(\boldsymbol{\sigma_i})$. To apply the Hoeffding inequality we need a bound on $\sup_{\boldsymbol{\sigma}} |g(\boldsymbol{\sigma})|$, which is derived for the case where $\mathcal{V} = \mathcal{H}_{\text{out}}$. Namely we assume that $\mathcal{V}$ is a set of all possible outputs of the algorithm (for a fixed $X_{m+u}$). Specifically, suppose that $\mathbf{v} \in \mathcal{V}$ is an output of the algorithm, $\mathbf{v} = U\boldsymbol{\alpha}$, and assume that $\|\boldsymbol{\alpha}\|_2 \leq \mu_1$.

By Definition 1, for all $\boldsymbol{\sigma}$, $\|\boldsymbol{\sigma}\|_2 \leq b \triangleq \sqrt{m+u}$. Let $\lambda_1 \leq \ldots \leq \lambda_k$ be the singular values of $U$ and $\mathbf{u}_1, \ldots, \mathbf{u}_k$ and $\mathbf{w}_1, \ldots, \mathbf{w}_k$ be their corresponding unit-length right and left singular vectors[10]. We have that

$$\sup_{\boldsymbol{\sigma}} |g(\boldsymbol{\sigma})| = \sup_{\|\boldsymbol{\sigma}\|_2 \leq b, \, \|\boldsymbol{\alpha}\|_2 \leq \mu_1} |\boldsymbol{\sigma}^T U \boldsymbol{\alpha}| = \sup_{\|\boldsymbol{\sigma}\|_2 \leq b, \, \|\boldsymbol{\alpha}\|_2 \leq \mu_1} \left| \boldsymbol{\sigma}^T \sum_{i=1}^{k} \lambda_i \mathbf{u}_i \mathbf{w}_i^T \boldsymbol{\alpha} \right| \leq b \mu_1 \lambda_k .$$

Applying the one-sided Hoeffding inequality on $n$ samples of $g(\boldsymbol{\sigma})$ we have, for any given $\delta$, that with probability of at least $1 - \delta$ over the random i.i.d. choice of the vectors $\boldsymbol{\sigma_1}, \ldots, \boldsymbol{\sigma_n}$,

$$R_{m+u}(\mathcal{V}) \leq \left( \frac{1}{m} + \frac{1}{u} \right) \cdot \left( \frac{1}{n} \sum_{i=1}^{n} \sup_{\boldsymbol{\alpha}:\|\boldsymbol{\alpha}\|_2 \leq \mu_1} \boldsymbol{\sigma}_i^T U \boldsymbol{\alpha} + \mu_1 \lambda_k \sqrt{m+u} \sqrt{\frac{2 \ln \frac{1}{\delta}}{n}} \right) . \quad (26)$$

To use the bound (26), the value of $\sup_{\boldsymbol{\alpha}:\|\boldsymbol{\alpha}\|_2 \leq \mu_1} \boldsymbol{\sigma}_i^T U \boldsymbol{\alpha}$ should be computed for each randomly drawn $\boldsymbol{\sigma}_i$. This computation is algorithm-dependent and in Section 6.3 we show how to compute it for the algorithm of Zhou et al. (2004).[11] In cases where we can compute the supremum exactly (as in vanilla ULRs; see below) we can also get a lower bound using the symmetric Hoeffding inequality.

# 6. Applications: Explicit Bounds for Specific Algorithms

In this section we exemplify the use of the Rademacher bounds (22), (25) and (26) to particular transductive algorithms. In Section 6.1 we instantiate the generic ULR bound (22) for the SGT algorithm of Joachims (2003). In Section 6.2 we instantiate kernel-ULR bound (25) for the algorithm of Belkin et al. (2004). Finally, in Section 6.3 we instantiate all three bounds (22), (25) and (26) for the algorithm of Zhou et al. (2004) and compare the resulting bounds numerically.

## 6.1 The Spectral Graph Transduction (SGT) Algorithm of Joachims (2003)

We start with a description of a simplified version of SGT that captures the essence of the algorithm.[12] Let $W$ be a symmetric $(m+u) \times (m+u)$ similarity matrix of the full-sample $X_{m+u}$. The $(i, j)$th entry of $W$ represents the similarity between $x_i$ and $x_j$. The matrix $W$ can be constructed in various ways, for example, it can be a $k$-nearest neighbors graph. In such graph each vertex represents example from the full sample $X_{m+u}$. There is an edge between a pair of vertices if one of the corresponding examples is among $k$ most similar examples to the other. The weights of the edges are proportional to the similarity of the adjacent vertices (points). The examples of commonly used measures of similarity are cosine similarity and RBF kernel. Let $D$ be a diagonal matrix, whose $(i, i)$th entry is the sum of the $i$th row in $W$. An unnormalized Laplacian of $W$ is $L = D - W$.

---

10. These vectors can be found from the singular value decomposition of $U$.

11. An application of this approach in induction seems to be very hard, if not impossible. For example, in the case of RBF kernel machines we will need to optimize over (typically) infinite-dimensional vectors in the feature space.

12. We omit a few heuristics that are optional in SGT. Their exclusion does not affect the error bound we derive.

Let $r \in \{1, \ldots, m + u - 1\}$ be fixed, $\{\lambda_i, \mathbf{v}_i\}_{i=1}^{m+u}$ be eigenvectors and eigenvalues of $L$ such that $0 = \lambda_1 \leq \ldots \leq \lambda_{m+u}$ and $\widetilde{L} = \sum_{i=2}^{r+1} i^2 \mathbf{v} \mathbf{v}^T$. Let $\boldsymbol{\tau} = (\tau_1, \ldots, \tau_{m+u})$ be a vector that specifies the given labels in $S_m$; that is, $\tau_i \in \{\pm 1\}$ for labeled points, and $\tau_i = 0$ otherwise. Let $c$ be a fixed constant and $\mathbf{1}$ be an $(m + u) \times 1$ vector whose entries are 1 and let $C$ be a diagonal matrix such that $C(i, i) = 1/m$ iff example $i$ is in the training set (and zero otherwise). The soft classification $\mathbf{h}^*$ produced by the SGT algorithm is the solution of the following optimization problem:

$$\min_{\mathbf{h} \in \mathbb{R}^{m+u}} \quad \mathbf{h}^T \widetilde{L} \mathbf{h} + c(\mathbf{h} - \vec{\tau})^T C(\mathbf{h} - \vec{\tau}) \tag{27}$$

$$s.t. \quad \mathbf{h}^T \mathbf{1} = 0, \qquad \mathbf{h}^T \mathbf{h} = m + u. \tag{28}$$

It is shown by Joachims (2003) that $\mathbf{h}^* = U \boldsymbol{\alpha}$, where $U$ is an $(m + u) \times r$ matrix whose columns are $\mathbf{v}_i$'s, $2 \leq i \leq r + 1$, and $\boldsymbol{\alpha}$ is an $r \times 1$ vector. While $\boldsymbol{\alpha}$ depends on both the training and test sets, the matrix $U$ depends only on the unlabeled full-sample. Substituting $\mathbf{h}^* = U \boldsymbol{\alpha}$ for the second constraint in (28) and using the orthonormality of the columns of $U$, we get $m + u = \mathbf{h}^{*T} \mathbf{h}^* = \boldsymbol{\alpha}^T U^T U \boldsymbol{\alpha} = \boldsymbol{\alpha}^T \boldsymbol{\alpha}$. Hence, $\|\boldsymbol{\alpha}\|_2 = \sqrt{m + u}$ and we can take $\mu_1 = \sqrt{m + u}$. Since $U$ is an $(m + u) \times r$ matrix with orthonormal columns, $\|U\|_{\text{Fro}}^2 = r$. We conclude from (22) the following bound on transductive Rademacher complexity of SGT

$$R_{m+u}(\mathcal{H}_{\text{out}}) \leq \sqrt{2r \left( \frac{1}{m} + \frac{1}{u} \right)}, \tag{29}$$

where $r$ is the number of non-zero eigenvalues of $L$. Notice that the bound (29) is oblivious to the magnitude of these eigenvalues. With the small value of $r$ the bound (29) is small, but, as shown by Joachims (2003) the test error of SGT is bad. If $r$ increases then the bound (29) increases but the test error improves. Joachims shows empirically that the smallest value of $r$ achieving nearly optimal test error is 40.

## 6.2 Kernel-ULR of the Algorithm of Belkin et al. (2004)

By defining the RKHS induced by the graph (unnormalized) Laplacian, as it was done by Herbster, Pontil, and Wainer (2005), and applying a generalized representer theorem of Schölkopf, Herbrich, and Smola (2001), we show that the algorithm of Belkin et al. (2004) has a kernel-ULR. Based on this kernel-ULR we derive an explicit risk bound for this. We also derive an explicit risk bound based on generic ULR. We show that the former (kernel) bound is tighter than the latter (generic) one. Finally, we compare our kernel bound with the risk bound of Belkin et al. (2004). The proofs of all lemmas in this section appear in Appendix G.

The algorithm of Belkin et al. (2004) is similar to the SGT algorithm, described in Section 6.1. Hence in this appendix we use the same notation as in the description of SGT (see Section 6.1). The algorithm of Belkin et al. is formulated as follows.

$$\min_{\mathbf{h} \in \mathbb{R}^{m+u}} \quad \mathbf{h}^T L \mathbf{h} + c(\mathbf{h} - \vec{\tau})^T C(\mathbf{h} - \vec{\tau}) \tag{30}$$

$$s.t. \qquad \mathbf{h}^T \mathbf{1} = 0 \tag{31}$$

The difference between (30)-(31) and (27)-(28) is in the constraint (28), which may change the resulting hard classification. Belkin et al. developed a stability-based error bound for the algorithm based on a connected graph. In the analysis that follows we also assume that the underlying graph is connected, but as shown at the end of this section, the argument can be also extended to unconnected graphs.

We represent a full-sample labeling as a vector in the Reproducing Kernel Hilbert Space (RKHS) associated with the graph Laplacian (as described by Herbster et al., 2005) and derive a transductive version of the generalized representer theorem of Schölkopf et al. (2001). Considering (30)-(31) we set $\mathcal{H} = \{\mathbf{h} \mid \mathbf{h}^T \mathbf{1} = 0, \ \mathbf{h} \in \mathbb{R}^{m+u}\}$. Let $\mathbf{h}_1, \mathbf{h}_2 \in \mathcal{H}$ be two soft classification vectors. We define their inner product as

$$\langle \mathbf{h}_1, \mathbf{h}_2 \rangle_L \triangleq \mathbf{h}_1^T L \mathbf{h}_2 \ . \tag{32}$$

We denote by $\mathcal{H}_L$ the set $\mathcal{H}$ along with the inner product (32). Let $\lambda_1, \ldots, \lambda_{m+u}$ be the eigenvalues of $L$ in the increasing order. Since $L$ is a Laplacian of the connected graph, $\lambda_1 = 0$ and for all $2 \leq i \leq m + u$, $\lambda_i \neq 0$. Let $\mathbf{u}_i$ be an eigenvector corresponding to $\lambda_i$. Since $L$ is symmetric, the vectors $\{\mathbf{u}_i\}_{i=1}^{m+u}$ are orthogonal. We assume also w.l.o.g. that the vectors $\{\mathbf{u}_i\}_{i=1}^{m+u}$ are orthonormal and $\mathbf{u}_1 = \frac{1}{\sqrt{m+u}}\mathbf{1}$. Let

$$U \triangleq \sum_{i=2}^{m+u} \frac{1}{\lambda_i} \mathbf{u}_i \mathbf{u}_i^T \ . \tag{33}$$

Note that the matrix $U$ depends only on the unlabeled full-sample.

**Lemma 7 (Herbster et al., 2005)** *The space $\mathcal{H}_L$ is an RKHS with a reproducing kernel matrix $U$.*

A consequence of Lemma 7 is that the algorithm (30)-(31) performs the regularization in the RKHS $\mathcal{H}_L$ with the regularization term $\|\mathbf{h}\|_L^2 = \mathbf{h}^T L \mathbf{h}$ (this fact was also noted by Herbster et al., 2005). The following transductive variant of the generalized representer theorem of Schölkopf et al. (2001) concludes the derivation of the kernel-ULR of the algorithm of Belkin et al. (2004).

**Lemma 8** *Let $\mathbf{h}^* \in \mathcal{H}$ be the solution of the optimization problem (30)-(31), and let $U$ be defined as above. Then, there exists $\boldsymbol{\alpha} \in \mathbb{R}^{m+u}$ such that $\mathbf{h}^* = U\boldsymbol{\alpha}$.*

**Remark 9** *We now consider the case of an unconnected graph. Let $t$ be the number of connected components in the underlying graph. Then the zero eigenvalue of the Laplacian $L$ has multiplicity $t$. Let $\mathbf{u}_1, \ldots, \mathbf{u}_t$ be the eigenvectors corresponding to the zero eigenvalue of $L$. Let $\mathbf{u}_{t+1}, \ldots, \mathbf{u}_{m+u}$ be the eigenvectors corresponding to non-zero eigenvalues* $\lambda_{t+1}, \ldots, \lambda_{m+u}$ of $L$. *We replace constraint (31) with $t$ constraints $\mathbf{h}^T \mathbf{u_i} = 0$ and define the kernel matrix as $U \triangleq \sum_{i=t+1}^{m+u} \frac{1}{\lambda_i} \mathbf{u}_i \mathbf{u}_i^T$. The rest of the analysis is the same as for the case of the connected graph.*

To obtain the explicit bounds on the transductive Rademacher complexity of the algorithm of Belkin et al. it remains to bound $\sqrt{\boldsymbol{\alpha}^T U \boldsymbol{\alpha}}$ and $\|\boldsymbol{\alpha}\|_2$. We start with bounding $\sqrt{\boldsymbol{\alpha}^T U \boldsymbol{\alpha}}$.

We substitute $\mathbf{h} = U\boldsymbol{\alpha}$ into (30)-(31). Since $\mathbf{u}_2, \ldots, \mathbf{u}_{m+u}$ are orthogonal to $\mathbf{u}_1 = \frac{1}{\sqrt{m+u}}\mathbf{1}$, we have that $\mathbf{h}^T \mathbf{1} = \boldsymbol{\alpha}^T U^T \mathbf{1} = \boldsymbol{\alpha}^T \sum_{i=2}^{m+u} \frac{1}{\lambda_i} \mathbf{u}_i \mathbf{u}_i^T \mathbf{1} = 0$. Moreover, we have that $\mathbf{h}^T L \mathbf{h} = \boldsymbol{\alpha}^T U^T L U \boldsymbol{\alpha} = \boldsymbol{\alpha}^T \left( I - \frac{1}{m+u} \mathbf{1} \cdot \mathbf{1}^T \right) U \boldsymbol{\alpha} = \boldsymbol{\alpha}^T U \boldsymbol{\alpha}$. Thus (30)-(31) is equivalent to solving

$$\min_{\boldsymbol{\alpha} \in \mathbb{R}^{m+u}} \boldsymbol{\alpha}^T U \boldsymbol{\alpha} + c(U\boldsymbol{\alpha} - \vec{\tau})^T C(U\boldsymbol{\alpha} - \vec{\tau}) \tag{34}$$

and outputting $\mathbf{h}^* = U\boldsymbol{\alpha}_{\text{out}}$, where $\boldsymbol{\alpha}_{\text{out}}$ is the solution of (34). Let $\mathbf{0}$ be the $(m+u) \times 1$ vector consisting of zeros. We have

$$\begin{aligned} \boldsymbol{\alpha}_{\text{out}}^T U \boldsymbol{\alpha}_{\text{out}} &\leq \boldsymbol{\alpha}_{\text{out}}^T U \boldsymbol{\alpha}_{\text{out}} + c(U\boldsymbol{\alpha}_{\text{out}} - \vec{\tau})^T C(U\boldsymbol{\alpha}_{\text{out}} - \vec{\tau}) \\ &\leq \mathbf{0}^T U \mathbf{0} + c(U\mathbf{0} - \vec{\tau})^T C(U\mathbf{0} - \vec{\tau}) = c \ . \end{aligned}$$

Thus

$$\sqrt{\boldsymbol{\alpha}_{\text{out}}^T U \boldsymbol{\alpha}_{\text{out}}} \leq \sqrt{c} \triangleq \mu_2 \ . \tag{35}$$

Let $\overline{\lambda}_1, \ldots, \overline{\lambda}_{m+u}$ be the eigenvalues of $U$, sorted in the increasing order. It follows from (33) that $\overline{\lambda}_1 = 0$ and for any $2 \leq i \leq m+u$, $\overline{\lambda}_i = \frac{1}{\lambda_{m+u-i+2}}$, where $\lambda_1, \ldots, \lambda_{m+u}$ are the eigenvalues of $L$ sorted in the increasing order.

We substitute the bound (35) into (25), and obtain that the kernel bound is

$$\sqrt{\frac{2c}{mu} \sum_{i=2}^{m+u} \overline{\lambda}_i} = \sqrt{\frac{2c}{mu} \sum_{i=2}^{m+u} \frac{1}{\lambda_i}} \ .$$

Suppose that[13] $\sum_{i=2}^{m+u} \frac{1}{\lambda_i} = O(m+u)$. We substitute the kernel bound into (15) and obtain that with probability at least $1 - \delta$ over the random training/test partition,

$$\mathcal{L}_u(\mathbf{h}) \leq \widehat{\mathcal{L}}_m^\gamma(\mathbf{h}) + O\left(\frac{1}{\sqrt{\min(m,u)}}\right) \ . \tag{36}$$

We briefly compare this bound with the risk bound for the algorithm (30)-(31) given by Belkin et al. (2004). Belkin et al. provide the following bound for their algorithm[14]. With probability of at least $1 - \delta$ over the random draw of $m$ training examples from $X_{m+u}$,

$$\mathcal{L}_{m+u}(\mathbf{h}) \leq \widehat{\mathcal{L}}_m^\gamma(\mathbf{h}) + O\left(\frac{1}{\sqrt{m}}\right) \ . \tag{37}$$

---

13. This assumption is not restricting since we can define the matrix $L$ and its spectrum after observing the unlabeled full-sample. Thus we can set $L$ in a way that this assumption will hold.
14. The original bound of Belkin et al. is in terms of squared loss. The equivalent bound in terms of 0/1 and margin loss can be obtained by the same derivation as in the paper of Belkin et al. (2004).

Similarly to what was done in Section 4.3, to bring the bounds to 'common denominator', we rewrite the bound (36) as

$$\mathcal{L}_u(\mathbf{h}) \leq \widehat{\mathcal{L}}_m^\gamma(\mathbf{h}) + \frac{u}{m+u} O\left(\frac{1}{\sqrt{\min(m,u)}}\right) \ . \tag{38}$$

If $m \ll u$ or $m \approx u$ then the bounds (37) and (38) have the same convergence rate. However if $m \gg u$ then the convergence rate of (38) (which is $O(1/(m\sqrt{u}))$) is much faster than the one of (37) (which is $O(1/\sqrt{m})$).

### 6.3 The Consistency Method of Zhou et al. (2004)

In this section we instantiate the bounds (22), (25) and (26) to the "consistency method" of Zhou et al. (2004) and provide their numerical comparison.

We start with a brief description of the Consistency Method (CM) algorithm of Zhou et al. (2004). The algorithm has a natural vanilla ULR (see definition at the beginning of Section 5), where the matrix $U$ is computed as follows. Let $W$ and $D$ be matrices as in SGT (see Section 6.1). Let $L \triangleq D^{-1/2} W D^{-1/2}$ and $\beta$ be a parameter in $(0,1)$. Then, $U \triangleq (1-\beta)(I - \beta L)^{-1}$ and the output of CM is $\mathbf{h} = U \cdot \boldsymbol{\alpha}$, where $\boldsymbol{\alpha}$ specifies the given labels. Consequently $\|\boldsymbol{\alpha}\|_2 = \sqrt{m}$. The following lemma, proven in Appendix H, provides a characterization of the eigenvalues of $U$:

**Lemma 9** *Let $\lambda_{\max}$ and $\lambda_{\min}$ be, respectively, the largest and smallest eigenvalues of $U$. Then $\lambda_{\max} = 1$ and $\lambda_{\min} > 0$.*

It follows from Lemma 9 that $U$ is a positive definite matrix and hence is also a kernel matrix. Therefore, the decomposition with the above $U$ is a kernel-ULR. To apply the kernel bound (25) we compute the bound $\mu_2$ on $\sqrt{\boldsymbol{\alpha}^T U \boldsymbol{\alpha}}$. By the Rayleigh-Ritz theorem (Horn & Johnson, 1990), we have that $\frac{\boldsymbol{\alpha}^T U \boldsymbol{\alpha}}{\boldsymbol{\alpha}^T \boldsymbol{\alpha}} \leq \lambda_{\max}$. Since by the definition of the vanilla ULR, $\boldsymbol{\alpha}^T \boldsymbol{\alpha} = m$, we obtain that $\sqrt{\boldsymbol{\alpha}^T U \boldsymbol{\alpha}} \leq \sqrt{\lambda_{\max} \boldsymbol{\alpha}^T \boldsymbol{\alpha}} = \sqrt{\lambda_{\max} m}$.

We obtained that $\mu_1 = \sqrt{m}$ and $\mu_2 = \sqrt{\lambda_{\max} m}$, where $\lambda_{\max}$ is the maximal eigenvalue of $U$. Since by Lemma 9 $\lambda_{\max} = 1$, for the CM algorithm the bound (22) is always tighter than (25).

It turns out that for CM, the exact value of the supremum in (26) can be analytically derived. Recall that the vectors $\boldsymbol{\alpha}$, which induce the CM hypothesis space for a particular $U$, have exactly $m$ components with values in $\{\pm 1\}$; the rest of the components are zeros. Let $\Psi$ be the set of all possible such $\boldsymbol{\alpha}$'s. Let $\mathbf{t}(\boldsymbol{\sigma}_i) = (t_1, \ldots, t_{m+u}) \triangleq \boldsymbol{\sigma}_i^T U \in \mathbb{R}^{1 \times (m+u)}$ and $|\mathbf{t}(\boldsymbol{\sigma}_i)| \triangleq (|t_1|, \ldots, |t_{m+u}|)$. Then, for any fixed $\boldsymbol{\sigma}_i$, $\sup_{\boldsymbol{\alpha} \in \Psi} \boldsymbol{\sigma}_i^T U \boldsymbol{\alpha}$ is the sum of the $m$ largest elements in $|\mathbf{t}(\boldsymbol{\sigma}_i)|$. This derivation holds for any vanilla ULR.

To demonstrate the Rademacher bounds discussed in this paper we present an empirical comparison of the bounds over two datasets (Voting, Pima) from the UCI repository[15]. For each dataset we took $m+u$ to be the size of the dataset (435 and 768 respectively) and we took $m$ to be $1/3$ of the full-sample size. The matrix $W$ is the 10-nearest neighbors graph computed with the cosine similarity metric. We applied the CM algorithm with $\beta = 0.5$. The Monte-Carlo bounds (both upper and lower) were computed with $\delta = 0.05$ and $n = 10^5$.

---

15. We also obtained similar results for several other UCI datasets.

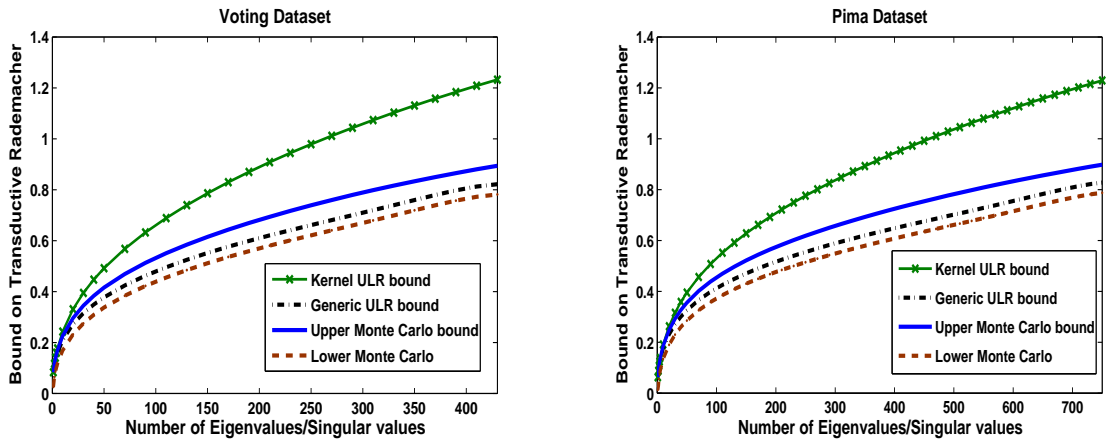

Figure 1: A comparison of transductive Rademacher bounds.

We compared upper and lower Mote-Carlo bounds with the generic ULR bound (22) and the kernel-ULR bound (25). The graphs in Figure 1 compare these four bounds for each of the datasets as a function of the number of non-zero eigenvalues of $U$. Specifically, each point $t$ on the $x$-axis corresponds to bounds computed with a matrix $U_t$ that approximates $U$ using only the smallest $t$ eigenvalues of $U$. In both examples the lower and upper Monte-Carlo bounds tightly "sandwich" the true Rademacher complexity. It is striking that generic-ULR bound is very close to the true Rademacher complexity. In principle, with our simple Monte-Carlo method we can approximate the true Rademacher complexity up to any desired accuracy (with high confidence) at the cost of drawing sufficiently many Rademacher vectors.

## 7. PAC-Bayesian Bound for Transductive Mixtures

In this section we adapt part of the results of Meir and Zhang (2003) to transduction. The proofs of all results presented in this section appear in Appendix I.

Let $\mathcal{B} = \{\mathbf{h}_i\}_{i=1}^{|\mathcal{B}|}$ be a finite set of *base-hypotheses*. The class $\mathcal{B}$ can be formed after observing the full-sample $X_{m+u}$, but before obtaining the training/test set partition and the labels. Let $\mathbf{q} = (q_1, \ldots, q_{|\mathcal{B}|}) \in \mathbb{R}^{|\mathcal{B}|}$ be a probability vector, i.e. $\sum_{i=1}^{|\mathcal{B}|} q_i = 1$ and $q_i \geq 0$ for all $1 \leq i \leq |\mathcal{B}|$. The vector $\mathbf{q}$ can be computed after observing training/test partition and the training labels. Our goal is to find the *"posterior"* vector $\mathbf{q}$ such that the *mixture hypothesis* $\widetilde{\mathbf{h}}_{\mathbf{q}} \overset{\triangle}{=} \sum_{i=1}^{|\mathcal{B}|} q_i \mathbf{h}_i$ minimizes $\mathcal{L}_u(\widetilde{\mathbf{h}}_{\mathbf{q}}) = \frac{1}{u} \sum_{j=m+1}^{m+u} \ell\left(\sum_{i=1}^{|\mathcal{B}|} q_i h_i(j), y_j\right)$.

In this section we derive a uniform risk bound for a set of $\mathbf{q}$'s. This bound depends on the KL-divergence (see the definition below) between $\mathbf{q}$ and the *"prior"* probability vector $\mathbf{p} \in \mathbb{R}^{|\mathcal{B}|}$, where the vector $\mathbf{p}$ is defined based only on the unlabeled full-sample. Thus our forthcoming bound (see Theorem 4) belongs to the family of PAC-Bayesian bounds (McAllester, 2003; Derbeko et al., 2004), which depend on prior and posterior information. Notice that our bound, is different from the PAC-Bayesian bounds for Gibbs classifiers that bound $\mathbf{E}_{\mathbf{h} \sim \mathcal{B}(\mathbf{q})} \mathcal{L}_u(\mathbf{h}) = \frac{1}{u} \sum_{j=m+1}^{m+u} \mathbf{E}_{\mathbf{h} \sim \mathcal{B}(\mathbf{q})} \ell(h(j), y_j)$, where $\mathbf{h} \sim \mathcal{B}(\mathbf{q})$ is a random draw of the base hypothesis from $\mathcal{B}$ according to distribution $\mathbf{q}$.

**Remark 10** *As by one of the reviewers noted, by Jensen inequality $\mathcal{L}_u(\widetilde{\mathbf{h}}_{\mathbf{q}}) \leq \mathbf{E}_{\mathbf{h} \sim \mathcal{B}(\mathbf{q})} \mathcal{L}_u(\mathbf{h})$. Hence any risk bound for transductive Gibbs classifier holds true also for transductive mixture classifier. Currently known risk bound for transductive Gibbs classifiers (Theorem 18 in the paper of Derbeko et al., 2004) diverges when $u \to \infty$. Our forthcoming risk bound (41) has no such deficiency.*

We assume that $\mathbf{q}$ belongs to domain $\Omega_{g,A} = \{\mathbf{q} \mid g(\mathbf{q}) \leq A\}$, where $g : \mathbb{R}^{|\mathcal{B}|} \to \mathbb{R}$ is a predefined function and $A \in \mathbb{R}$ is a constant. The domain $\Omega_{g,A}$ and the set $\mathcal{B}$ induce the class $\widetilde{\mathcal{B}}_{g,A}$ of all possible mixtures $\widetilde{\mathbf{h}}_{\mathbf{q}}$. Recalling that $Q \triangleq (1/m + 1/u)$, $S \triangleq \frac{m+u}{(m+u-0.5)(1-0.5/\max(m,u))}$ and $c_0 = \sqrt{32 \ln(4e)/3} < 5.05$, we apply Theorem 2 with $\mathcal{H}_{\text{out}} \triangleq \widetilde{\mathcal{B}}_{g,A}$ and obtain that with probability of at least $1 - \delta$ over the training/test partition of $X_{m+u}$, for all $\widetilde{\mathbf{h}}_{\mathbf{q}} \in \widetilde{\mathcal{B}}_{g,A}$,

$$\mathcal{L}_u(\widetilde{\mathbf{h}}_{\mathbf{q}}) \leq \widehat{\mathcal{L}}_m^{\gamma}(\widetilde{\mathbf{h}}_{\mathbf{q}}) + \frac{R_{m+u}(\widetilde{\mathcal{B}}_{g,A})}{\gamma} + c_0 Q \sqrt{\min(m,u)} + \sqrt{\frac{S}{2} Q \ln \frac{1}{\delta}}. \qquad (39)$$

Let $Q_1 \triangleq \sqrt{\frac{S}{2} Q \left( \ln(1/\delta) + 2 \ln \log_s \left( s\tilde{g}(\mathbf{q})/g_0 \right) \right)}$. It is straightforward to apply the technique used in the proof of Theorem 10 of Meir and Zhang (2003) and obtain the following bound, which eliminates the dependence on $A$.

**Corollary 2** *Let $g_0 > 0$, $s > 1$ and $\tilde{g}(\mathbf{q}) = s \max(g(\mathbf{q}), g_0)$. For any fixed $g$ and $\gamma > 0$, with probability of at least $1 - \delta$ over the training/test set partition, for all[16] $\mathbf{h}_{\mathbf{q}}$,*

$$\mathcal{L}_u(\widetilde{\mathbf{h}}_{\mathbf{q}}) \leq \widehat{\mathcal{L}}_m^{\gamma}(\widetilde{\mathbf{h}}_{\mathbf{q}}) + \frac{R_{m+u}(\widetilde{\mathcal{B}}_{g,\tilde{g}(\mathbf{q})})}{\gamma} + c_0 Q \sqrt{\min(m,u)} + Q_1 . \qquad (40)$$

We now instantiate Corollary 2 for $g(\mathbf{q})$ being the KL-divergence and derive a PAC-Bayesian bound. Let $g(\mathbf{q}) \triangleq D(\mathbf{q}\|\mathbf{p}) = \sum_{i=1}^{|\mathcal{B}|} q_i \ln \left( \frac{q_i}{p_i} \right)$ be KL-divergence between $\mathbf{p}$ and $\mathbf{q}$. Adopting Lemma 11 of Meir and Zhang (2003) to the transductive Rademacher variables, defined in (1), we obtain the following bound.

**Theorem 4** *Let $g_0 > 0$, $s > 1$, $\gamma > 0$. Let $\mathbf{p}$ and $\mathbf{q}$ be any prior and posterior distribution over $\mathcal{B}$, respectively. Set $g(\mathbf{q}) \triangleq D(\mathbf{q}\|\mathbf{p})$ and $\tilde{g}(\mathbf{q}) \triangleq s \max(g(\mathbf{q}), g_0)$. Then, with probability of at least $1 - \delta$ over the training/test set partition, for all $\mathbf{h}_{\mathbf{q}}$,*

$$\mathcal{L}_u(\widetilde{\mathbf{h}}_{\mathbf{q}}) \leq \widehat{\mathcal{L}}_m^{\gamma}(\widetilde{\mathbf{h}}_{\mathbf{q}}) + \frac{Q}{\gamma} \sqrt{2\tilde{g}(\mathbf{q}) \sup_{\mathbf{h} \in \mathcal{B}} \|\mathbf{h}\|_2^2} + c_0 Q \sqrt{\min(m,u)} + Q_1 . \qquad (41)$$

Theorem 4 is a PAC-Bayesian result, where the prior $\mathbf{p}$ can depend on $X_{m+u}$ and the posterior can be optimized adaptively, based also on $S_m$. As our general bound (15), the bound (41) has the convergence rate of $O(1/\sqrt{\min(m,u)})$. The bound (41) is syntactically similar to inductive PAC-Bayesian bound for mixture hypothesis (see Theorem 10 and Lemma 11 in the paper of Meir & Zhang, 2003), having similar convergence rate of $O(1/\sqrt{m})$. However the conceptual difference between inductive and transductive bounds is that in transduction we can define the prior vector $\mathbf{p}$ after observing the unlabeled full-sample and in induction we should define $\mathbf{p}$ before observing any data.

---

16. In the bound (40) the meaning of $R_{m+u}(\widetilde{B}_{g,\tilde{g}(\mathbf{q})})$ is as follows: for any $\mathbf{q}$, let $A = \tilde{g}(\mathbf{q})$ and $R_{m+u}(\widetilde{\mathcal{B}}_{g,\tilde{g}(\mathbf{q})}) \triangleq R_{m+u}(\widetilde{\mathcal{B}}_{g,A})$.

## 8. Concluding Remarks

We studied the use of Rademacher complexity analysis in the transductive setting. Our results include the first general Rademacher bound for soft classification algorithms, the unlabeled-labeled representation (ULR) technique for bounding the Rademacher complexity of any transductive algorithm and a bound for Bayesian mixtures. We demonstrated the usefulness of these results and, in particular, the effectiveness of our ULR framework for deriving error bounds for several advanced transductive algorithms.

It would be nice to further improve our bounds using, for example, the local Rademacher approach of Bartlett, Bousquet, and Mendelson (2005). However, we believe that the main advantage of these transductive bounds is the possibility of selecting a hypothesis space based on a full-sample. A clever data-dependent choice of this space should provide sufficient flexibility to achieve a low training error with low Rademacher complexity. In our opinion this opportunity can be explored and exploited much further. In particular, it would be interesting to develop an efficient procedure for the choice of hypothesis space if the learner knows the properties of the underlying distribution (e.g., if the clustering assumption holds).

This work opens up new avenues for future research. For example, it would be interesting to optimize the matrix $U$ in the ULR explicitly (to fit the data) under a constraint of low Rademacher complexity. Also, it would be nice to find "low-Rademacher" approximations of particular $U$ matrices. The PAC-Bayesian bound for mixture algorithms motivates the development and use of transductive mixtures, an area that has yet to be investigated. Finally, it would be interesting to utilize our bounds in model selection process.

### Acknowledgments

We are grateful to anonymous reviewers for their insightful comments. We also thank Yair Wiener and Nati Srebro for fruitful discussions. Dmitry Pechony was supported in part by the IST Programme of the European Community, under the PASCAL Network of Excellence, IST-2002-506778.

### Appendix A. Proof of Lemma 1

The proof is based on the technique used in the proof of Lemma 5 in the paper of Meir and Zhang (2003). Let $\boldsymbol{\sigma} = (\sigma_1, \ldots, \sigma_{m+u})^T$ be the Rademacher random variables of $R_{m+u}(\mathcal{V}, p_1)$ and $\boldsymbol{\tau} = (\tau_1, \ldots, \tau_{m+u})^T$ be the Rademacher random variables of $R_{m+u}(\mathcal{V}, p_2)$. For any real-valued function $g(\mathbf{v})$, for any $n \in I_1^{m+u}$ and any $\mathbf{v}' \in \mathcal{V}$,

$$\sup_{\mathbf{v} \in \mathcal{V}} [g(\mathbf{v})] = \mathbf{E}_{\tau_n} \left\{ \tau_n v_n' + \sup_{\mathbf{v} \in \mathcal{V}} [g(\mathbf{v})] \, \middle| \, \tau_n \neq 0 \right\} \leq \mathbf{E}_{\tau_n} \left\{ \sup_{\mathbf{v} \in \mathcal{V}} [\tau_n v_n + g(\mathbf{v})] \, \middle| \, \tau_n \neq 0 \right\} \ . \tag{42}$$

We use the abbreviation $\tau_1^s \stackrel{\triangle}{=} \tau_1, \ldots, \tau_s$. We apply (42) with a fixed $\tau_1^{n-1}$ and $g(\mathbf{v}) \stackrel{\triangle}{=} f(\mathbf{v}) + \sum_{i=1}^{n-1} \tau_i v_i$, and obtain that

$$\sup_{\mathbf{v} \in \mathcal{V}} \left[ \sum_{i=1}^{n-1} \tau_i v_i + f(\mathbf{v}) \right] \leq \mathbf{E}_{\tau_n} \left\{ \sup_{\mathbf{v} \in \mathcal{V}} \left[ \sum_{i=1}^{n} \tau_i v_i + f(\mathbf{v}) \right] \, \middle| \, \tau_n \neq 0 \right\} \ . \tag{43}$$

To complete the proof of the lemma, we prove a more general claim: for any real-valued function $f(\mathbf{v})$, for any $0 \leq n \leq m+u$,

$$\mathbf{E}_{\boldsymbol{\sigma}} \left\{ \sup_{\mathbf{v} \in \mathcal{V}} \left[ \sum_{i=1}^{n} \sigma_i v_i + f(\mathbf{v}) \right] \right\} \leq \mathbf{E}_{\boldsymbol{\tau}} \left\{ \sup_{\mathbf{v} \in \mathcal{V}} \left[ \sum_{i=1}^{n} \tau_i v_i + f(\mathbf{v}) \right] \right\} . \qquad (44)$$

The proof is by induction on $n$. The claim trivially holds for $n = 0$ (in this case (44) holds with equality). Suppose the claim holds for all $k < n$ and all functions $f(\mathbf{v})$. We use the abbreviation $\sigma_1^s \triangleq \sigma_1, \ldots, \sigma_s$. For any function $f'(\mathbf{v})$ have

$$\mathbf{E}_{\sigma_1^n} \sup_{\mathbf{v} \in \mathcal{V}} \left[ \sum_{i=1}^{n} \sigma_i v_i + f'(\mathbf{v}) \right]$$

$$= 2p_1 \left\{ \frac{1}{2} \mathbf{E}_{\sigma_1^{n-1}} \sup_{\mathbf{v} \in \mathcal{V}} \left[ \sum_{i=1}^{n-1} \sigma_i v_i + v_n + f'(\mathbf{v}) \right] + \frac{1}{2} \mathbf{E}_{\sigma_1^{n-1}} \sup_{\mathbf{v} \in \mathcal{V}} \left[ \sum_{i=1}^{n-1} \sigma_i v_i - v_n + f'(\mathbf{v}) \right] \right\}$$

$$+ (1 - 2p_1) \, \mathbf{E}_{\sigma_1^{n-1}} \sup_{\mathbf{v} \in \mathcal{V}} \left[ \sum_{i=1}^{n-1} \sigma_i v_i + f'(\mathbf{v}) \right]$$

$$\leq 2p_1 \left\{ \frac{1}{2} \mathbf{E}_{\tau_1^{n-1}} \sup_{\mathbf{v} \in \mathcal{V}} \left[ \sum_{i=1}^{n-1} \tau_i v_i + v_n + f'(\mathbf{v}) \right] \right. \qquad (45)$$

$$\left. + \frac{1}{2} \mathbf{E}_{\tau_1^{n-1}} \sup_{\mathbf{v} \in \mathcal{V}} \left[ \sum_{i=1}^{n-1} \tau_i v_i - v_n + f'(\mathbf{v}) \right] \right\} + (1 - 2p_1) \, \mathbf{E}_{\tau_1^{n-1}} \sup_{\mathbf{v} \in \mathcal{V}} \left[ \sum_{i=1}^{n-1} \tau_i v_i + f'(\mathbf{v}) \right]$$

$$= \mathbf{E}_{\tau_1^{n-1}} \left\{ 2p_1 \mathbf{E}_{\tau_n} \left\{ \sup_{\mathbf{v} \in \mathcal{V}} \left[ \sum_{i=1}^{n} \tau_i v_i + f'(\mathbf{v}) \right] \; \middle| \; \tau_n \neq 0 \right\} + (1 - 2p_1) \sup_{\mathbf{v} \in \mathcal{V}} \left[ \sum_{i=1}^{n-1} \tau_i v_i + f'(\mathbf{v}) \right] \right\}$$

$$= \mathbf{E}_{\tau_1^{n-1}} \left\{ 2p_1 \left( \mathbf{E}_{\tau_n} \left\{ \sup_{\mathbf{v} \in \mathcal{V}} \left[ \sum_{i=1}^{n} \tau_i v_i + f'(\mathbf{v}) \right] \; \middle| \; \tau_n \neq 0 \right\} - \sup_{\mathbf{v} \in \mathcal{V}} \left[ \sum_{i=1}^{n-1} \tau_i v_i + f'(\mathbf{v}) \right] \right) \right.$$

$$\left. + \sup_{\mathbf{v} \in \mathcal{V}} \left[ \sum_{i=1}^{n-1} \tau_i v_i + f'(\mathbf{v}) \right] \right\}$$

$$\leq \mathbf{E}_{\tau_1^{n-1}} \left\{ 2p_2 \left( \mathbf{E}_{\tau_n} \left\{ \sup_{\mathbf{v} \in \mathcal{V}} \left[ \sum_{i=1}^{n} \tau_i v_i + f'(\mathbf{v}) \right] \; \middle| \; \tau_n \neq 0 \right\} - \sup_{\mathbf{v} \in \mathcal{V}} \left[ \sum_{i=1}^{n-1} \tau_i v_i + f'(\mathbf{v}) \right] \right) \right.$$

$$\left. + \sup_{\mathbf{v} \in \mathcal{V}} \left[ \sum_{i=1}^{n-1} \tau_i v_i + f'(\mathbf{v}) \right] \right\} \qquad (46)$$

$$= \mathbf{E}_{\tau_1^{n-1}} \left\{ 2p_2 \mathbf{E}_{\tau_n} \left\{ \sup_{\mathbf{v} \in \mathcal{V}} \left[ \sum_{i=1}^{n} \tau_i v_i + f'(\mathbf{v}) \right] \; \middle| \; \tau_n \neq 0 \right\} \right.$$

$$\left. + (1 - 2p_2) \sup_{\mathbf{v} \in \mathcal{V}} \left[ \sum_{i=1}^{n-1} \tau_i v_i + f'(\mathbf{v}) \right] \right\}$$

$$= \mathbf{E}_{\tau_1^n} \sup_{\mathbf{v} \in \mathcal{V}} \left[ \sum_{i=1}^{n} \tau_i v_i + f'(\mathbf{v}) \right] .$$

The inequality (45) follow from the inductive hypothesis, applied thrice with $f(\mathbf{v}) = v_n + f'(\mathbf{v})$, $f(\mathbf{v}) = -v_n + f'(\mathbf{v})$ and $f(\mathbf{v}) = f'(\mathbf{v})$. The inequality (46) follows from (43) and the fact that $p_1 < p_2$.

## Appendix B. Proof of Lemma 2

We require the following standard definitions and facts about martingales.[17] Let $\mathbf{B}_1^n \triangleq (B_1, \ldots, B_n)$ be a sequence of random variables and $\mathbf{b}_1^n \triangleq (b_1, \ldots, b_n)$ be their respective values. The sequence $\mathbf{W}_0^n \triangleq (W_0, W_1, \ldots, W_n)$ is called a *martingale* w.r.t. the *underlying sequence* $\mathbf{B}_1^n$ if for any $i \in I_1^n$, $W_i$ is a function of $\mathbf{B}_1^i$ and $\mathbf{E}_{B_i}\{W_i | \mathbf{B}_1^{i-1}\} = W_{i-1}$.

Let $f(\mathbf{X}_1^n) \triangleq f(X_1, \ldots, X_n)$ be an arbitrary function of $n$ (possibly dependent) random variables. Let $W_0 \triangleq \mathbf{E}_{\mathbf{X}_1^n}\{f(\mathbf{X}_1^n)\}$ and $W_i \triangleq \mathbf{E}_{\mathbf{X}_1^n}\{f(\mathbf{X}_1^n) | \mathbf{X}_1^i\}$ for any $i \in I_1^n$. An elementary fact is that $\mathbf{W}_0^n$ is a martingale w.r.t. the underlying sequence $\mathbf{X}_1^n$. Thus we can obtain a martingale from any function of (possibly dependent) random variables. This routine of obtaining a martingale from an arbitrary function is called *Doob's martingale process*. By the definition of $W_n$ we have $W_n = \mathbf{E}_{\mathbf{X}_1^n}\{f(\mathbf{X}_1^n) | \mathbf{X}_1^n\} = f(\mathbf{X}_1^n)$. Consequently, to bound the deviation of $f(\mathbf{X}_1^n)$ from its mean it is sufficient to bound the difference $W_n - W_0$. A fundamental inequality, providing such a bound, is McDiarmid's inequality (McDiarmid, 1989).

**Lemma 10 (McDiarmid, 1989, Corollary 6.10)** *Let $\mathbf{W}_0^n$ be a martingale w.r.t. $\mathbf{B}_1^n$. Let $\mathbf{b}_1^n = (b_1, \ldots, b_n)$ be the vector of possible values of the random variables $B_1, \ldots, B_n$. Let*

$$r_i(\mathbf{b}_1^{i-1}) \triangleq \sup_{b_i}\left\{W_i \ : \ \mathbf{B}_1^{i-1} = \mathbf{b}_1^{i-1}, B_i = b_i\right\} - \inf_{b_i}\left\{W_i \ : \ \mathbf{B}_1^{i-1} = \mathbf{b}_1^{i-1}, B_i = b_i\right\} \ .$$

*Let $r^2(\mathbf{b}_1^n) \triangleq \sum_{i=1}^n (r_i(\mathbf{b}_1^{i-1}))^2$ and $\widehat{r}^2 \triangleq \sup_{\mathbf{b}_1^n} r^2(\mathbf{b}_1^n)$. Then,*

$$\mathbf{P}_{\mathbf{B}_1^n}\{W_n - W_0 > \epsilon\} < \exp\left(-\frac{2\epsilon^2}{\widehat{r}^2}\right) \ . \tag{47}$$

The inequality (47) is an improved version of the Hoeffding-Azuma inequality (Hoeffding, 1963; Azuma, 1967).

The proof of Lemma 2 is inspired by McDiarmid's proof of the bounded difference inequality for permutation graphs (McDiarmid, 1998, Section 3). Let $\mathbf{W}_0^{m+u}$ be a martingale obtained from $f(\mathbf{Z})$ by Doob's martingale process, namely $W_0 \triangleq \mathbf{E}_{\mathbf{Z}_1^{m+u}}\{f(\mathbf{Z}_1^{m+u})\}$ and $W_i \triangleq \mathbf{E}_{\mathbf{Z}_1^{m+u}}\{f(\mathbf{Z}_1^{m+u}) | \mathbf{Z}_1^i\}$. We compute the upper bound on $\widehat{r}^2$ and apply Lemma 10.

Fix $i$, $i \in I_1^m$. Let $\boldsymbol{\pi}_1^{m+u} = \pi_1, \ldots, \pi_{m+u}$ be a specific permutation of $I_1^{m+u}$ and $\pi_i' \in \{\pi_{i+1}, \ldots, \pi_{m+u}\}$. Let $p_1 \triangleq \mathbf{P}_{j \sim I_{i+1}^{m+u}}\{j \in I_{i+1}^m\} = \frac{m-i}{m+u-i}$ and $p_2 \triangleq \mathbf{P}_{j \sim I_{i+1}^{m+u}}\{j \in I_{m+1}^{m+u}\} =$

---

17. See, e.g., Chapter 12 of Grimmett and Stirzaker (1995), and Section 9.1 of Devroye et al. (1996) for more details.

$1 - p_1 = \frac{u}{m+u-i}$. We have

$$
\begin{aligned}
r_i(\boldsymbol{\pi}_1^{i-1}) &= \sup_{\pi_i} \left\{ W_i \; : \; \mathbf{B}_1^{i-1} = \boldsymbol{\pi}_1^{i-1}, B_i = \pi_i \right\} - \inf_{\pi_i} \left\{ W_i \; : \; \mathbf{B}_1^{i-1} = \boldsymbol{\pi}_1^{i-1}, B_i = \pi_i \right\} \\
&= \sup_{\pi_i, \pi_i'} \left\{ \mathbf{E}_{\mathbf{Z}} \left\{ f(\mathbf{Z}) \mid \mathbf{Z}_1^{i-1} = \boldsymbol{\pi}_1^{i-1}, Z_i = \pi_i \right\} - \mathbf{E}_{\mathbf{Z}} \left\{ f(\mathbf{Z}) \mid \mathbf{Z}_1^{i-1} = \boldsymbol{\pi}_1^{i-1}, Z_i = \pi_i' \right\} \right\} \\
&= \sup_{\pi_i, \pi_i'} \left\{ \mathbf{E}_{j \sim I_{i+1}^{m+u}} \mathbf{E}_{\mathbf{Z}} \left\{ f(\mathbf{Z}) \mid \mathbf{Z}_1^{i-1} = \boldsymbol{\pi}_1^{i-1}, Z_i = \pi_i, Z_j = \pi_i' \right\} \right. \\
&\qquad\qquad \left. - \mathbf{E}_{j \sim I_{i+1}^{m+u}} \mathbf{E}_{\mathbf{Z}} \left\{ f(\mathbf{Z}^{ij}) \mid \mathbf{Z}_1^{i-1} = \boldsymbol{\pi}_1^{i-1}, Z_i = \pi_i, Z_j = \pi_i' \right\} \right\} \\
&= \sup_{\pi_i, \pi_i'} \left\{ \mathbf{E}_{j \sim I_{i+1}^{m+u}} \mathbf{E}_{\mathbf{Z}} \left\{ f(\mathbf{Z}) - f(\mathbf{Z}^{ij}) \mid \mathbf{Z}_1^{i-1} = \boldsymbol{\pi}_1^{i-1}, Z_i = \pi_i, Z_j = \pi_i' \right\} \right\} \qquad (48) \\
&= \sup_{\pi_i, \pi_i'} \left\{ p_1 \cdot \mathbf{E}_{\mathbf{Z}, j \sim I_{i+1}^{m}} \left\{ f(\mathbf{Z}) - f(\mathbf{Z}^{ij}) \mid \mathbf{Z}_1^{i-1} = \boldsymbol{\pi}_1^{i-1}, Z_i = \pi_i, Z_j = \pi_i' \right\} \right. \qquad (49) \\
&\qquad\qquad \left. + p_2 \cdot \mathbf{E}_{\mathbf{Z}, j \sim I_{m+1}^{m+u}} \left\{ f(\mathbf{Z}) - f(\mathbf{Z}^{ij}) \mid \mathbf{Z}_1^{i-1} = \boldsymbol{\pi}_1^{i-1}, Z_i = \pi_i, Z_j = \pi_i' \right\} \right\}
\end{aligned}
$$

Since $f(\mathbf{Z})$ is $(m, u)$-permutation symmetric function, the expectation in (49) is zero. Therefore,

$$
\begin{aligned}
r_i(\boldsymbol{\pi}_1^{i-1}) &= \sup_{\pi_i, \pi_i'} \left\{ p_2 \cdot \mathbf{E}_{\mathbf{Z}, j \sim I_{m+1}^{m+u}} \left\{ f(\mathbf{Z}) - f(\mathbf{Z}^{ij}) \mid \mathbf{Z}_1^{i-1} = \boldsymbol{\pi}_1^{i-1}, Z_i = \pi_i, Z_j = \pi_i' \right\} \right\} \\
&\leq \frac{u\beta}{m+u-i} \; .
\end{aligned}
$$

Since $f(\mathbf{Z})$ is $(m, u)$-permutation symmetric, it also follows from (48) that for $i \in I_{m+1}^{m+u}$, $r_i(\boldsymbol{\pi}_1^{i-1}) = 0$. It can be verified that for any $j > 1/2$, $\frac{1}{j^2} \leq \int_{j-1/2}^{j+1/2} \frac{1}{t^2} \mathrm{d}t$, and therefore,

$$
\begin{aligned}
\widehat{r}^2 &= \sup_{\boldsymbol{\pi}_1^{m+u}} \sum_{i=1}^{m+u} \left( r_i(\boldsymbol{\pi}_1^{i-1}) \right)^2 \leq \sum_{i=1}^{m} \left( \frac{u\beta}{m+u-i} \right)^2 = u^2 \beta^2 \sum_{j=u}^{m+u-1} \frac{1}{j^2} \\
&\leq u^2 \beta^2 \int_{u-1/2}^{m+u-1/2} \frac{1}{t^2} \mathrm{d}t = \frac{mu^2 \beta^2}{(u-1/2)(m+u-1/2)} \; . \qquad (50)
\end{aligned}
$$

By applying Lemma 10 with the bound (50) we obtain

$$
\mathbf{P}_{\mathbf{Z}} \left\{ f(\mathbf{Z}) - \mathbf{E}_{\mathbf{Z}} \left\{ f(\mathbf{Z}) \right\} \geq \epsilon \right\} \leq \exp \left( -\frac{2\epsilon^2 (u-1/2)(m+u-1/2)}{mu^2 \beta^2} \right) \; . \qquad (51)
$$

The entire derivation is symmetric in $m$ and $u$. Therefore, we also have

$$
\mathbf{P}_{\mathbf{Z}} \left\{ f(\mathbf{Z}) - \mathbf{E}_{\mathbf{Z}} \left\{ f(\mathbf{Z}) \right\} \geq \epsilon \right\} \leq \exp \left( -\frac{2\epsilon^2 (m-1/2)(m+u-1/2)}{m^2 u \beta^2} \right) \; . \qquad (52)
$$

By taking the tightest bound from (51) and (52) we obtain the statement of the lemma.

## Appendix C. Proof of Lemma 3

We consider the following algorithm[18] (named `RANDPERM`) for drawing the first $m$ elements $\{Z_i\}_{i=1}^m$ of the random permutation $\mathbf{Z}$ of $I_1^{m+u}$:

---
1: Let $Z_i = i$ for any $i \in I_1^{m+u}$.
2: **for** $i = 1$ to $m$ **do**
3:     Draw $d_i$ uniformly from $I_i^{m+u}$.
4:     Swap the values of $Z_i$ and $Z_{d_i}$.
5: **end for**
---

**Algorithm 1:** `RANDPERM` - draw the first $m$ elements of the random permutation of $m + u$ elements.

The algorithm `RANDPERM` is an abridged version of the procedure of drawing a random permutation of $n$ elements by drawing $n-1$ non-identically distributed independent random variables, presented in Section 5 of the paper of Talagrand (1995) (which according to Talagrand is due to Maurey, 1979).

**Lemma 11** *The algorithm* `RANDPERM` *performs a uniform draw of the first $m$ elements $Z_1, \ldots, Z_m$ of the random permutation $\mathbf{Z}$.*

**Proof:** The proof is by induction on $m$. If $m = 1$, then a single random variable $d_1$ is uniformly drawn among $I_{m+u}$, and therefore, $Z_1$ has a uniform distribution over $I_1^{m+u}$. Let $\mathbf{d}_1^m \triangleq d_1, \ldots, d_m$. Suppose the claim holds for all $m_1 < m$. For any two possible values $\boldsymbol{\pi}_1^m \triangleq \pi_1, \ldots, \pi_m$ and $\boldsymbol{\pi}_1'^m \triangleq \pi_1', \ldots, \pi_m'$ of $Z_1, \ldots, Z_m$, we have

$$
\begin{aligned}
\mathbf{P}_{\mathbf{d}_1^m}\{\mathbf{Z}_1^m = \boldsymbol{\pi}_1^m\} &= \mathbf{P}_{\mathbf{d}_1^{m-1}}\{\mathbf{Z}_1^{m-1} = \boldsymbol{\pi}_1^{m-1}\} \cdot \mathbf{P}_{d_m}\{Z_m = \pi_m \mid \mathbf{Z}_1^{m-1} = \boldsymbol{\pi}_1^{m-1}\} \\
&= \mathbf{P}_{\mathbf{d}_1^{m-1}}\{\mathbf{Z}_1^{m-1} = \boldsymbol{\pi}_1'^{m-1}\} \cdot \frac{1}{u+1} \\
&= \mathbf{P}_{\mathbf{d}_1^{m-1}}\{\mathbf{Z}_1^{m-1} = \boldsymbol{\pi}_1'^{m-1}\} \cdot \mathbf{P}_{d_m}\{Z_m = \pi_m' \mid \mathbf{Z}_1^{m-1} = \boldsymbol{\pi}_1'^{m-1}\} \\
&= \mathbf{P}_{\mathbf{d}_1^m}\{\mathbf{Z}_1^m = \boldsymbol{\pi}_1'^m\} \ .
\end{aligned}
\tag{53}
$$

The equality (53) follows from the inductive assumption and the definition of $d_m$. $\qquad\square$

Consider any $(m, u)$-permutation symmetric function $f = f(\mathbf{Z})$ over random permutations $\mathbf{Z}$. Using the algorithm `RANDPERM` we can represent any random permutation $\mathbf{Z}$ as a function $g(\mathbf{d})$ of $m$ independent random variables. The value of the function $g(\mathbf{d})$ is the output of the algorithm `RANDPERM` operated with the values of random draws given by $\mathbf{d}$. The next lemma relates the Lipschitz constant of the function $f(g(\mathbf{d}))$ to the Lipschitz constant of $f(\mathbf{Z})$:

---

18. Another algorithm for generating random permutation from independent draws was presented in Appendix B of Lanckriet et al. (2004). This algorithm draws a random permutation by means of drawing $m + u$ independent random variables. Since we only deal with $(m, u)$-permutation symmetric functions, we are only interested in the first $m$ elements of the random permutation. The algorithm of Lanckriet et al. needs $m + u$ draws of independent random variables to define the above $m$ elements. The algorithm `RANDPERM`, presented in this section, needs only $m$ draws. If we use the algorithm of Lanckriet et al. instead of `RANDPERM`, the forthcoming bound (55) would have the term $m + u$ instead of $m$. This change, in turn, would result in a non-convergent risk bound being derived using our techniques.

**Lemma 12** *Let $f(\mathbf{Z})$ be an $(m, u)$-permutation symmetric function of random permutation $\mathbf{Z}$. Suppose that for all $i \in I_1^m$, $j \in I_{m+1}^{m+u}$, $|f(\mathbf{Z}) - f(\mathbf{Z}^{ij})| \leq \beta$. Let $d_i'$ be an independent draw of the random variable $d_i$. Then for any $i \in I_1^m$,*

$$|f(g(d_1, \ldots, d_{i-1}, d_i, d_{i+1}, \ldots, d_m)) - f(g(d_1, \ldots, d_{i-1}, d_i', d_{i+1}, \ldots, d_m))| \leq \beta \ . \tag{54}$$

**Proof:** The values of $\mathbf{d} \triangleq (d_1, \ldots, d_i, \ldots, d_m)$ and $\mathbf{d}' \triangleq (d_1, \ldots, d_i', \ldots, d_m)$ induce, respectively, the first $m$ values[19] $\mathbf{Z}_1^m = \{Z_1, \ldots, Z_m\}$ and $\mathbf{Z}_1'^m = \{Z_1', \ldots, Z_m'\}$ of the two dependent permutations of $I_1^{m+u}$. Since $f$ is $(m, u)$-permutation symmetric, its value is uniquely determined by the value of $\mathbf{Z}_1^m$. We prove that the change of $d_i$ by $d_i'$ results in a change of a single element in $\mathbf{Z}_1^m$. Combined with the property of $|f(\mathbf{Z}) - f(\mathbf{Z}^{ij})| \leq \beta$, this will conclude the proof of (54).

We refer to $\mathbf{d}$ and $\mathbf{d}'$ as, respectively, 'old' and 'new' draws. Consider the operation of RANDPERM with the draws $\mathbf{d}$ and $\mathbf{d}'$. Let $\pi_i, \pi_{d_i}$ and $\pi_{d_i'}$ be the values of, respectively, $Z_i$, $Z_{d_i}$ and $Z_{d_i'}$ just *before* the $i$th iteration of RANDPERM. Note that $d_i \geq i$ and $d_i' \geq i$. In the old permutation, *after* the $i$th iteration $Z_i = \pi_{d_i}$, $Z_{d_i} = \pi_i$ and $Z_{d_i'} = \pi_{d_i'}$. In the new permutation, *after* the $i$th iteration $Z_i = \pi_{d_i'}$, $Z_{d_i} = \pi_{d_i}$ and $Z_{d_i'} = \pi_i$. After the $i$th iteration of RANDPERM the value of $Z_i$ remains intact. However the values of $Z_{d_i}$ and $Z_{d_i'}$ may change. In particular the values of $\pi_{d_i}$ and $\pi_i$ may be among $Z_{i+1}, \ldots, Z_m$ at the end of the run of RANDPERM. We have four cases:

**Case 1** If $\pi_{d_i'} \notin \mathbf{Z}_1^m$ and $\pi_i \notin \mathbf{Z}_1^m$ then $\pi_{d_i} \notin \mathbf{Z}_1'^m$, $\pi_i \notin \mathbf{Z}_1'^m$ and $\mathbf{Z}_1'^m = \mathbf{Z}_1^m \backslash \{\pi_{d_i}\} \cup \{\pi_{d_i'}\}$.

**Case 2** If $\pi_{d_i'} \in \mathbf{Z}_1^m$ and $\pi_i \in \mathbf{Z}_1^m$ then $\pi_{d_i} \in \mathbf{Z}_1'^m$, $\pi_i \in \mathbf{Z}_1'^m$ and $\mathbf{Z}_1'^m = \mathbf{Z}_1^m$.

**Case 3** If $\pi_i \in \mathbf{Z}_1^m$ and $\pi_{d_i'} \notin \mathbf{Z}_1^m$ then $\pi_{d_i} \in \mathbf{Z}_1'^m$, $\pi_i \notin \mathbf{Z}_1'^m$ and $\mathbf{Z}_1'^m = \mathbf{Z}_1^m \backslash \{\pi_i\} \cup \{\pi_{d_i'}\}$.

**Case 4** If $\pi_{d_i'} \in \mathbf{Z}_1^m$ and $\pi_i \notin \mathbf{Z}_1^m$ then $\pi_i \in \mathbf{Z}_1'^m$, $\pi_{d_i} \notin \mathbf{Z}_1'^m$ and $\mathbf{Z}_1'^m = \mathbf{Z}_1^m \backslash \{\pi_{d_i}\} \cup \{\pi_i\}$.

$\square$

We apply a bounded difference inequality of McDiarmid (1989) to $f(g(\mathbf{d}))$ and obtain

$$\mathbf{P_d}\left\{f(g(\mathbf{d})) - \mathbf{E_d}\left\{f(g(\mathbf{d}))\right\} \geq \epsilon\right\} \leq \exp\left(-\frac{2\epsilon^2}{\beta^2 m}\right) \ . \tag{55}$$

Since $f(\mathbf{Z})$ is a $(m, u)$-permutation symmetric, it follows from (55) that

$$\mathbf{P_Z}\left\{f(\mathbf{Z}) - \mathbf{E_Z}\left\{f(\mathbf{Z})\right\} \geq \epsilon\right\} \leq \exp\left(-\frac{2\epsilon^2}{\beta^2 m}\right) \ . \tag{56}$$

Since the entire derivation is symmetric in $m$ and $u$ we also have

$$\mathbf{P_Z}\left\{f(\mathbf{Z}) - \mathbf{E_Z}\left\{f(\mathbf{Z})\right\} \geq \epsilon\right\} \leq \exp\left(-\frac{2\epsilon^2}{\beta^2 u}\right) \ . \tag{57}$$

The proof of Lemma 3 is completed by taking the minimum of the bounds (56) and (57).

---

19. For notational convenience in this section, we refer to $\mathbf{Z}_1^m$ as a set of values and not as a vector of values (as is done in other sections).

## Appendix D. Proof of Claims in Lemma 4

**Proof of Claim 1.** Note that $N_1$ and $N_2$ are random variables whose distribution is induced by the distribution of $\tilde{\boldsymbol{\sigma}}$. We have by (9) that

$$\tilde{R}_{m+u}(\mathcal{V}) = \mathbf{E}_{N_1,N_2}\mathbf{E}_{\tilde{\boldsymbol{\sigma}}\sim\text{Rad}(N_1,N_2)}\sup_{\mathbf{v}\in\mathcal{V}}\sum_{i=1}^{m+u}(\tilde{\sigma}_{i,1}+\tilde{\sigma}_{i,2})\,v(i) = \mathbf{E}_{N_1,N_2}s(N_1,N_2)\ .$$

**Proof of Claim 2.** By the definitions of $\mathbf{H}_k$ and $\mathbf{T}_k$ (appearing at the start of Section 4.1), for any $N_1, N_2 \in I_1^{m+u}$ we have

$$\mathbf{E}_{\mathbf{Z},\mathbf{Z}'}\sup_{\mathbf{v}\in\mathcal{V}}\Big[\mathbf{T}_{N_1}\{\mathbf{v}(\mathbf{Z})\} - \mathbf{T}_{N_2}\{\mathbf{v}(\mathbf{Z}')\} + \mathbf{H}_{N_2}\{\mathbf{v}(\mathbf{Z}')\} - \mathbf{H}_{N_1}\{\mathbf{v}(\mathbf{Z})\}\Big] =$$

$$\mathbf{E}_{\mathbf{Z},\mathbf{Z}'}\sup_{\mathbf{v}\in\mathcal{V}}\underbrace{\left[\frac{1}{u}\sum_{i=N_1+1}^{m+u}v(Z_i) - \frac{1}{u}\sum_{i=N_2+1}^{m+u}v(Z_i') + \frac{1}{m}\sum_{i=1}^{N_2}v(Z_i') - \frac{1}{m}\sum_{i=1}^{N_1}v(Z_i)\right]}_{\triangleq r(\mathbf{v},\mathbf{Z},\mathbf{Z}',N_1,N_2)}. \quad (58)$$

The values of $N_1$ and $N_2$, and the distribution of $\mathbf{Z}$ and $\mathbf{Z}'$, with respect to which we take the expectation in (58), induce a distribution of assignments of coefficients $\left\{\frac{1}{m}, -\frac{1}{m}, \frac{1}{u}, -\frac{1}{u}\right\}$ to the components of $\mathbf{v}$. For any $N_1, N_2$ and realizations of $\mathbf{Z}$ and $\mathbf{Z}'$, each component $v(i)$, $i \in I_1^{m+u}$, is assigned to exactly two coefficients, one for each of the two permutations ($\mathbf{Z}$ and $\mathbf{Z}'$). Let $\mathbf{a} \triangleq (a_1, \ldots, a_{m+u})$, where $a_i \triangleq (a_{i,1}, a_{i,2})$ is a pair of coefficients. For any $i \in I_1^{m+u}$, the pair $(a_{i,1}, a_{i,2})$ takes the values of the coefficients of $v(i)$, where the first component is induced by the realization $\mathbf{Z}$ (i.e., $a_{i,1}$ is either $-\frac{1}{m}$ or $\frac{1}{u}$) and the second component by the realization of $\mathbf{Z}'$ (i.e., $a_{i,2}$ is either $\frac{1}{m}$ or $-\frac{1}{u}$).

Let $\mathbf{A}(N_1, N_2)$ be the distribution of vectors $\mathbf{a}$, induced by the distribution of $\mathbf{Z}$ and $\mathbf{Z}'$, for particular $N_1, N_2$. Using this definition we can write

$$(58) = \mathbf{E}_{\mathbf{a}\sim\mathbf{A}(N_1,N_2)}\sup_{\mathbf{v}\in\mathcal{V}}\left[\sum_{i=1}^{m+u}(a_{i,1}+a_{i,2})v(i)\right]\ . \quad (59)$$

Let $\text{Par}(\text{k})$ be the uniform distribution over partitions of $m+u$ elements into two subsets, of $k$ and $m + u - k$ elements, respectively. Clearly, $\text{Par}(k)$ is a uniform distribution over $\binom{m+u}{k}$ elements. The distribution of the random vector $(a_{1,1}, a_{2,1}, \ldots, a_{m+u,1})$ of the first elements of pairs in $\mathbf{a}$ is equivalent to $\text{Par}(N_1)$. That is, this vector is obtained by taking the first $N_1$ indices of the realization of $\mathbf{Z}$ and assigning $-\frac{1}{m}$ to the corresponding components. The other components are assigned to $\frac{1}{u}$. Similarly, the distribution of the random vector $(a_{1,2}, a_{2,2}, \ldots, a_{m+u,2})$ is equivalent to $\text{Par}(N_2)$. Therefore, the distribution $\mathbf{A}(N_1, N_2)$ of the entire vector $\mathbf{a}$ is equivalent to the product distribution of $\text{Par}(N_1)$ and $\text{Par}(N_2)$, which is a uniform distribution over $\binom{m+u}{N_1}\cdot\binom{m+u}{N_2}$ elements, where each element is a pair of independent permutations.

We show that the distributions $\text{Rad}(N_1, N_2)$ and $\mathbf{A}(N_1, N_2)$ are identical. Given $N_1$ and $N_2$ and setting $\omega = (m+u)^2$, the probability of drawing a specific realization of $\tilde{\boldsymbol{\sigma}}$ (satisfying

$n_1 + n_2 = N_1$ and $n_2 + n_3 = N_2$) is

$$\left(\frac{m^2}{\omega}\right)^{n_2}\left(\frac{mu}{\omega}\right)^{N_1-n_2}\left(\frac{mu}{\omega}\right)^{N_2-n_2}\left(\frac{u^2}{\omega}\right)^{m+u-N_1-N_2+n_2} = \frac{m^{N_1+N_2}u^{2(m+u)-N_1-N_2}}{(m+u)^{2(m+u)}} \quad . \quad (60)$$

Since (60) is independent of the $n_i$'s, the distribution $\mathtt{Rad}(N_1, N_2)$ is uniform over all possible Rademacher assignments satisfying the constraints $N_1$ and $N_2$. It is easy to see that the support size of $\mathtt{Rad}(N_1, N_2)$ is the same as the support size of $\mathtt{A}(N_1, N_2)$. Moreover, the support sets of these distributions are identical; hence these distributions are identical. Therefore, it follows from (59) that

$$(58) = \mathbf{E}_{\tilde{\boldsymbol{\sigma}}\sim\mathtt{Rad}(N_1,N_2)}\left\{\sup_{\mathbf{v}\in\mathcal{V}}\left[\sum_{i=1}^{m+u}(\tilde{\sigma}_{i,1}+\tilde{\sigma}_{i,2})v(i)\right]\right\} = s(N_1, N_2) \quad .$$

It is easy to see that $\mathbf{E}_{\tilde{\boldsymbol{\sigma}}}N_1 = \mathbf{E}_{\tilde{\boldsymbol{\sigma}}}\{n_1+n_2\} = m$ and that $\mathbf{E}_{\tilde{\boldsymbol{\sigma}}}N_2 = \mathbf{E}_{\tilde{\boldsymbol{\sigma}}}\{n_2+n_3\} = m$. Since $\mathbf{E}_{\mathbf{Z}}\{\psi(\mathbf{Z})\}$ is (58) with $N_1 = m$ and $N_2 = m$, we have

$$\mathbf{E}_{\mathbf{Z}}\{\psi(\mathbf{Z})\} = \mathbf{E}_{\tilde{\boldsymbol{\sigma}}\sim\mathtt{Rad}(m,m)}\left\{\sup_{\mathbf{v}\in\mathcal{V}}\left[\sum_{i=1}^{m+u}(\tilde{\sigma}_{i,1}+\tilde{\sigma}_{i,2})\,v(i)\right]\right\} = s\left(\mathbf{E}_{\tilde{\boldsymbol{\sigma}}}N_1, \mathbf{E}_{\tilde{\boldsymbol{\sigma}}}N_2\right).$$

**Proof of Claim 3.**

We bound the differences $|s(N_1, N_2) - s(N_1', N_2)|$ and $|s(N_1, N_2) - s(N_1, N_2')|$ for any $1 \leq N_1, N_2, N_1', N_2' \leq m + u$. Suppose w.l.o.g. that $N_1' \leq N_1$. Recalling the definition of $r(\cdot)$ in (58) we have

$$s(N_1, N_2) = \mathbf{E}_{\mathbf{Z},\mathbf{Z}'}\sup_{\mathbf{v}\in\mathcal{V}}\left[r(\mathbf{v}, \mathbf{Z}, \mathbf{Z}', N_1, N_2)\right]$$

$$s(N_1', N_2) = \mathbf{E}_{\mathbf{Z},\mathbf{Z}'}\sup_{\mathbf{v}\in\mathcal{V}}\left[r(\mathbf{v}, \mathbf{Z}, \mathbf{Z}', N_1, N_2) + \left(\frac{1}{u}+\frac{1}{m}\right)\sum_{i=N_1'+1}^{N_1}v(Z_i)\right]. \quad (61)$$

The expressions under the supremums in $s(N_1, N_2)$ and $s(N_1', N_2)$ differ only in the two terms in (61). Therefore, for any $N_1$ and $N_1'$,

$$\left|s(N_1, N_2) - s(N_1', N_2)\right| \leq B_{\max}\left|N_1 - N_1'\right|\left(\frac{1}{u}+\frac{1}{m}\right) \quad . \quad (62)$$

Similarly we have that for any $N_2$ and $N_2'$,

$$\left|s(N_1, N_2) - s(N_1, N_2')\right| \leq B_{\max}\left|N_2 - N_2'\right|\left(\frac{1}{u}+\frac{1}{m}\right) \quad . \quad (63)$$

We use the following Bernstein-type concentration inequality (see Devroye et al., 1996, Problem 8.3) for the binomial random variable $X \sim \mathtt{Bin}(p, n)$: $\mathbf{P}_X\{|X - \mathbf{E}X| > t\} < 2\exp\left(-\frac{3t^2}{8np}\right)$. Abbreviate $Q \triangleq \frac{1}{m}+\frac{1}{u}$. Noting that $N_1, N_2 \sim \mathtt{Bin}\left(\frac{m}{m+u}, m+u\right)$, we use (62), (63) and the Bernstein-type inequality (applied with $n \triangleq m + u$ and $p \triangleq \frac{m}{m+u}$) to obtain

$$
\begin{aligned}
& \mathbf{P}_{N_1,N_2}\left\{|s(N_1,N_2) - s(\mathbf{E}_{\tilde{\boldsymbol{\sigma}}}\{N_1\}, \mathbf{E}_{\tilde{\boldsymbol{\sigma}}}\{N_2\})| \geq \epsilon\right\} \\
\leq\ & \mathbf{P}_{N_1,N_2}\left\{|s(N_1,N_2) - s(N_1, \mathbf{E}_{\tilde{\boldsymbol{\sigma}}}N_2)| + |s(N_1, \mathbf{E}_{\tilde{\boldsymbol{\sigma}}}N_2) - s(\mathbf{E}_{\tilde{\boldsymbol{\sigma}}}N_1, \mathbf{E}_{\tilde{\boldsymbol{\sigma}}}N_2)| \geq \epsilon\right\} \\
\leq\ & \mathbf{P}_{N_1,N_2}\left\{|s(N_1,N_2) - s(N_1, \mathbf{E}_{\tilde{\boldsymbol{\sigma}}}N_2)| \geq \frac{\epsilon}{2}\right\} \\
& + \mathbf{P}_{N_1,N_2}\left\{|s(N_1, \mathbf{E}_{\tilde{\boldsymbol{\sigma}}}N_2) - s(\mathbf{E}_{\tilde{\boldsymbol{\sigma}}}N_1, \mathbf{E}_{\tilde{\boldsymbol{\sigma}}}N_2)| \geq \frac{\epsilon}{2}\right\} \\
\leq\ & \mathbf{P}_{N_2}\left\{|N_2 - \mathbf{E}_{\tilde{\boldsymbol{\sigma}}}N_2| B_{\max}Q \geq \frac{\epsilon}{2}\right\} + \mathbf{P}_{N_1}\left\{|N_1 - \mathbf{E}_{\tilde{\boldsymbol{\sigma}}}N_1| B_{\max}Q \geq \frac{\epsilon}{2}\right\} \\
\leq\ & 4\exp\left(-\frac{3\epsilon^2}{32(m+u)\frac{m}{m+u}B_{\max}^2 Q^2}\right) = 4\exp\left(-\frac{3\epsilon^2}{32mB_{\max}^2 Q^2}\right)\ .
\end{aligned}
$$

Next we use the following fact (see Devroye et al., 1996, Problem 12.1): if a nonnegative random variable $X$ satisfies $\mathbf{P}\{X > t\} \leq c \cdot \exp(-kt^2)$ for some $c \geq 1$ and $k > 0$, then $\mathbf{E}X \leq \sqrt{\ln(ce)/k}$. Using this fact, along with $c \triangleq 4$ and $k \triangleq 3/(32mQ^2)$, we have

$$
\begin{aligned}
|\mathbf{E}_{N_1,N_2}\{s(N_1,N_2)\} - s(\mathbf{E}_{\tilde{\boldsymbol{\sigma}}}N_1, \mathbf{E}_{\tilde{\boldsymbol{\sigma}}}N_2)| & \leq \mathbf{E}_{N_1,N_2}|s(N_1,N_2) - s(\mathbf{E}_{\tilde{\boldsymbol{\sigma}}}N_1, \mathbf{E}_{\tilde{\boldsymbol{\sigma}}}N_2)| \\
& \leq \sqrt{\frac{32\ln(4e)}{3}mB_{\max}^2\left(\frac{1}{u} + \frac{1}{m}\right)^2}\ .
\end{aligned}
$$

## Appendix E. Proof of Lemma 5

The proof is a straightforward extension of the proof of Lemma 5 from Meir and Zhang (2003) and is also similar to the proof of our Lemma 1 in Appendix A. We prove a stronger claim: if for all $i \in I_1^{m+u}$ and $\mathbf{v}, \mathbf{v}' \in \mathcal{V}$, $|f(v_i) - f(v_i')| \leq |g(v_i) - g(v_i')|$, then for any function $\tilde{c} \colon \mathbb{R}^{m+u} \to \mathbb{R}$.

$$
\mathbf{E}_{\boldsymbol{\sigma}}\sup_{\mathbf{v}\in\mathcal{V}}\left[\sum_{i=1}^{m+u}\sigma_i f(v_i) + \tilde{c}(\mathbf{v})\right] \leq \mathbf{E}_{\boldsymbol{\sigma}}\sup_{\mathbf{v}\in\mathcal{V}}\left[\sum_{i=1}^{m+u}\sigma_i g(v_i) + \tilde{c}(\mathbf{v})\right]\ .
$$

We use the abbreviation $\sigma_1^n \triangleq \sigma_1, \ldots, \sigma_n$. The proof is by induction on $n$, such that $0 \leq n \leq m + u$. The lemma trivially holds for $n = 0$. Suppose the lemma holds for $n - 1$. In other words, for any function $\tilde{c}(\mathbf{v})$,

$$
\mathbf{E}_{\sigma_1^{n-1}}\sup_{\mathbf{v}\in\mathcal{V}}\left[\tilde{c}(\mathbf{v}) + \sum_{i=1}^{n-1}\sigma_i f(v_i)\right] \leq \mathbf{E}_{\sigma_1^{n-1}}\sup_{\mathbf{v}\in\mathcal{V}}\left[\tilde{c}(\mathbf{v}) + \sum_{i=1}^{n-1}\sigma_i g(v_i)\right]\ .
$$

Let $p \triangleq \frac{mu}{(m+u)^2}$. We have

$$
A \triangleq \mathbf{E}_{\sigma_1^n}\sup_{\mathbf{v}\in\mathcal{V}}\left[c(\mathbf{v}) + \sum_{i=1}^{n}\sigma_i f(v_i)\right] = \mathbf{E}_{\sigma_n}\mathbf{E}_{\sigma_1^{n-1}}\sup_{\mathbf{v}\in\mathcal{V}}\left[c(\mathbf{v}) + \sum_{i=1}^{n}\sigma_i f(v_i)\right] \quad (64)
$$

$$= p\mathbf{E}_{\sigma_1^{n-1}}\left\{\sup_{\mathbf{v}\in\mathcal{V}}\left[c(\mathbf{v})+\sum_{i=1}^{n-1}\sigma_i f(v_i)+f(v_n)\right]+\sup_{\mathbf{v}\in\mathcal{V}}\left[c(\mathbf{v})+\sum_{i=1}^{n-1}\sigma_i f(v_i)-f(v_n)\right]\right\} \quad (65)$$

$$+(1-2p)\mathbf{E}_{\sigma_1^{n-1}}\sup_{\mathbf{v}\in\mathcal{V}}\left[c(\mathbf{v})+\sum_{i=1}^{n-1}\sigma_i f(v_i)\right]. \quad (66)$$

We apply the inductive hypothesis three times: on the first and second summands in (65) with $\tilde{c}(\mathbf{v})\triangleq c(\mathbf{v})+f(v_n)$ and $\tilde{c}(\mathbf{v})\triangleq c(\mathbf{v})-f(v_n)$, respectively, and on (66) with $\tilde{c}(\mathbf{v})\triangleq c(\mathbf{v})$. We obtain

$$A \leq \underbrace{p\mathbf{E}_{\sigma_1^{n-1}}\left\{\sup_{\mathbf{v}\in\mathcal{V}}\left[c(\mathbf{v})+\sum_{i=1}^{n-1}\sigma_i g(v_i)+f(v_n)\right]+\sup_{\mathbf{v}\in\mathcal{V}}\left[c(\mathbf{v})+\sum_{i=1}^{n-1}\sigma_i g(v_i)-f(v_n)\right]\right\}}_{\triangleq B}$$

$$\underbrace{+(1-2p)\mathbf{E}_{\sigma_1^{n-1}}\sup_{\mathbf{v}\in\mathcal{V}}\left[c(\mathbf{v})+\sum_{i=1}^{n-1}\sigma_i g(v_i)\right]}_{\triangleq C}.$$

The expression $B$ can be written as follows.

$$B = p\mathbf{E}_{\sigma_1^{n-1}}\left\{\sup_{\mathbf{v}\in\mathcal{V}}\left[c(\mathbf{v})+\sum_{i=1}^{n-1}\sigma_i g(v_i)+f(v_n)\right]+\sup_{\mathbf{v}'\in\mathcal{V}}\left[c(\mathbf{v}')+\sum_{i=1}^{n-1}\sigma_i g(v_i')-f(v_n')\right]\right\}$$

$$= p\mathbf{E}_{\sigma_1^{n-1}}\sup_{\mathbf{v},\mathbf{v}'\in\mathcal{V}}\left[c(\mathbf{v})+c(\mathbf{v}')+\sum_{i=1}^{n-1}\left[\sigma_i(g(v_i)+g(v_i'))\right]+(f(v_n)-f(v_n'))\right]$$

$$= p\mathbf{E}_{\sigma_1^{n-1}}\sup_{\mathbf{v},\mathbf{v}'\in\mathcal{V}}\left[c(\mathbf{v})+c(\mathbf{v}')+\sum_{i=1}^{n-1}\left[\sigma_i(g(v_i)+g(v_i'))\right]+\left|f(v_n)-f(v_n')\right|\right]. \quad (67)$$

The equality (67) holds since the expression $c(\mathbf{v})+c(\mathbf{v}')+\sum_{i=1}^{n-1}\sigma_i(g(v_i)+g(v_i'))$ is symmetric in $\mathbf{v}$ and $\mathbf{v}'$. Thus, if $f(\mathbf{v})<f(\mathbf{v}')$ then we can exchange the values of $\mathbf{v}$ and $\mathbf{v}'$ and this will increase the value of the expression under the supremum. Since $|f(v_n)-f(v_n')| \leq |g(v_n)-g(v_n')|$ we have

$$B \leq p\mathbf{E}_{\sigma_1^{n-1}}\sup_{\mathbf{v},\mathbf{v}'\in\mathcal{V}}\left[c(\mathbf{v})+c(\mathbf{v}')+\sum_{i=1}^{n-1}\left[\sigma_i(g(v_i)+g(v_i'))\right]+|g(v_n)-g(v_n')|\right]$$

$$= p\mathbf{E}_{\sigma_1^{n-1}}\sup_{\mathbf{v},\mathbf{v}'\in\mathcal{V}}\left[c(\mathbf{v})+c(\mathbf{v}')+\sum_{i=1}^{n-1}\left[\sigma_i(g(v_i)+g(v_i'))\right]+(g(v_n)-g(v_n'))\right]$$

$$= p\mathbf{E}_{\sigma_1^{n-1}}\left\{\sup_{\mathbf{v}\in\mathcal{V}}\left[c(\mathbf{v})+\sum_{i=1}^{n-1}\sigma_i g(v_i)+g(v_n)\right]+\sup_{\mathbf{v}\in\mathcal{V}}\left[c(\mathbf{v})+\sum_{i=1}^{n-1}\sigma_i g(v_i)-g(v_n)\right]\right\}\triangleq D.$$

Therefore, using the reverse argument of (64)-(66),

$$A \leq C + D = \mathbf{E}_{\sigma_1^n}\sup_{\mathbf{v}\in\mathcal{V}}\left[c(\mathbf{v})+\sum_{i=1}^{n}\sigma_i g(v_i)\right].$$

## Appendix F. Proof of Lemma 6

Let $c \in \mathbb{R}$, $U \stackrel{\triangle}{=} c \cdot I$. If $c = 0$, then the soft classification generated by $\mathcal{A}$ is a constant zero. In this case, for any $\mathbf{h}$ generated by $\mathcal{A}$, we have $\widehat{\mathcal{L}}_m(\mathbf{h}) = 1$ and the lemma holds.

Suppose $c \neq 0$. Then

$$\boldsymbol{\alpha} = \frac{1}{c} \cdot \mathbf{h} \ . \tag{68}$$

Since the $(m + u) \times (m + u)$ matrix $U$ has $m + u$ singular values, each one is precisely $c$, by (22) the Rademacher complexity of the trivial ULR is bounded by

$$\mu_1 \sqrt{\frac{2}{mu}(m + u)c^2} = c\mu_1 \sqrt{2\left(\frac{1}{m} + \frac{1}{u}\right)} \ . \tag{69}$$

We assume w.l.o.g. that the training points have indices from 1 to $m$. Let $A = \{i \in I_1^m \mid y_i h(i) > 0 \text{ and } |h(i)| > \gamma\}$ be a set of indices of training examples with zero margin loss. Let $B = \{i \in I_1^m \mid |h(i)| \in [-\gamma, \gamma]\}$ and $C = \{i \in I_1^m \mid y_i h(i) < 0 \text{ and } |h(i)| > \gamma\}$. By (68) and the definition of the sets $A$, $C$, for any $i \in A \cup C$, $|\alpha_i| > \frac{\gamma}{c}$. Similarly, for any $i \in B$, $|\alpha_i| = \frac{|h(i)|}{c}$. We obtain that the bound (69) is at least

$$c\sqrt{(|A| + |C|)\frac{\gamma^2}{c^2} + \sum_{i \in B} \frac{h(i)^2}{c^2}} \sqrt{\frac{1}{m}} \ .$$

Therefore, the risk bound (15) is bounded from below by

$$\widehat{\mathcal{L}}_m^\gamma(\mathbf{h}) + \frac{1}{\gamma} \sqrt{(|A| + |C|)\gamma^2 + \sum_{i \in B} h(i)^2} \cdot \sqrt{\frac{2}{m}} \quad \geq$$

$$\frac{\sum_{i \in B}(1 - |h(i)|/\gamma) + |C|}{m} + \sqrt{|A| + |C| + \sum_{i \in B} \frac{h(i)^2}{\gamma^2}} \cdot \sqrt{\frac{2}{m}} \quad =$$

$$\frac{|B| + |C| - \sum_{i \in B} r_i}{m} + \sqrt{|A| + |C| + \sum_{i \in B} r_i^2} \cdot \sqrt{\frac{2}{m}} \quad =$$

$$\frac{m - |A| - \sum_{i \in B} r_i}{m} + \sqrt{|A| + |C| + \sum_{i \in B} r_i^2} \cdot \sqrt{\frac{2}{m}} \quad \stackrel{\triangle}{=} \quad D \ ,$$

where $r_i = \frac{|h(i)|}{\gamma}$. We prove that $D \geq 1$. Equivalently, it is sufficient to prove that for $r_{i_1}, \ldots, r_{i_{|B|}} \in [0, 1]^{|B|}$ it holds that

$$f\left(r_{i_1}, \ldots, r_{i_{|B|}}\right) = \frac{(|A| + \sum_{i \in B} r_i)^2}{|A| + |C| + \sum_{i \in B} r_i^2} \leq m \ .$$

We claim that the stronger statement holds:

$$f\left(r_{i_1}, \ldots, r_{i_{|B|}}\right) = \frac{(|A| + |C| + \sum_{i \in B} r_i)^2}{|A| + |C| + \sum_{i \in B} r_i^2} \leq m \ . \tag{70}$$

To prove (70) we use the Cauchy-Schwarz inequality, stating that for any two vectors $\mathbf{a}, \mathbf{b} \in \mathbb{R}^m$, $\langle \mathbf{a}, \mathbf{b} \rangle \leq \|\mathbf{a}\|_2 \cdot \|\mathbf{b}\|_2$. We set $b_i = 1$ for all $i \in I_1^m$. The vector $\mathbf{a}$ is set as follows: $a_i \triangleq r_i$ if $i \in B$ and $a_i = 1$ otherwise. By this definition of $\mathbf{a}$ and $\mathbf{b}$, we have that $\langle \mathbf{a}, \mathbf{b} \rangle \geq 0$ and thus $(\langle \mathbf{a}, \mathbf{b} \rangle)^2 \leq \|\mathbf{a}\|_2^2 \cdot \|\mathbf{b}\|_2^2$. The application of this inequality with the defined vectors $\mathbf{a}$ and $\mathbf{b}$ results in the inequality (70).

## Appendix G. Proofs from Section 6.2

**Proof of Lemma 7:** Let $\mathbf{e}_i$ be an $(m + u) \times 1$ vector whose $i$th entry equals 1 and other entries are zero. According to the definition of RKHS, we need to show that for any $1 \leq i \leq m + u$, $h(i) = \langle U(i, \cdot), \mathbf{h} \rangle_L$. We have

$$
\begin{aligned}
\langle U(i, \cdot), \mathbf{h} \rangle_L &= U(i, \cdot) L \mathbf{h} = \mathbf{e}_i U L \mathbf{h} \\
&= \mathbf{e}_i^T \left( \sum_{i=2}^{m+u} \frac{1}{\lambda_i} \mathbf{u}_i \mathbf{u}_i^T \right) \left( \sum_{i=1}^{m+u} \lambda_i \mathbf{u}_i \mathbf{u}_i^T \right) \mathbf{h} = \mathbf{e}_i^T \left( \sum_{i=2}^{m+u} \mathbf{u}_i \mathbf{u}_i^T \right) \mathbf{h} \\
&= \mathbf{e}_i^T (I - \mathbf{u}_1 \mathbf{u}_1^T) \mathbf{h} = \mathbf{e}_i^T \left( I - \frac{1}{m+u} \mathbf{1} \cdot \mathbf{1}^T \right) \mathbf{h} = h(i) \;\; .
\end{aligned}
$$

$\square$

**Lemma 13** *For any $1 \leq i \leq m + u$, $U(i, \cdot) \in \mathcal{H}_L$.*

**Proof:** Since $L$ is a Laplacian matrix, $\mathbf{u}_1 = \mathbf{1}$. Since the vectors $\{\mathbf{u}_i\}_{i=1}^{m+u}$ are orthonormal and $\mathbf{u}_1 = \mathbf{1}$, we have $U \cdot \mathbf{1} = \left( \sum_{i=2}^{m+u} \frac{1}{\lambda_i} \mathbf{u}_i \mathbf{u}_i^T \right) \mathbf{1} = 0$. Therefore, for any $1 \leq i \leq m + u$, $U(i, \cdot) \cdot \mathbf{1} = 0$. $\square$

**Proof of Lemma 8:** Let $\|\mathbf{h}\|_L = \sqrt{\langle \mathbf{h}, \mathbf{h} \rangle_L} \triangleq \sqrt{\mathbf{h}^T L \mathbf{h}}$ be a norm in $\mathcal{G}_L$. The optimization problem (30)-(31) can be stated in the following form:

$$
\min_{\mathbf{h} \in \mathcal{H}_L} \quad \|\mathbf{h}\|_L^2 + c(\mathbf{h} - \vec{\tau})^T C(\mathbf{h} - \vec{\tau}) \;\; . \tag{71}
$$

Let $\mathcal{U} \subseteq \mathcal{H}_L$ be a vector space spanned by the vectors $\{U(i, \cdot)\}_{i=1}^{m+u}$. Let $\mathbf{h}_\parallel \triangleq \sum_{i=1}^{m+u} \alpha_i U(i, \cdot)$ be a projection of $\mathbf{h}$ onto $\mathcal{U}$. For any $1 \leq i \leq m + u$, $\alpha_i = \frac{\langle \mathbf{h}, U(i, \cdot) \rangle_L}{\|U(i, \cdot)\|_L}$. Let $\mathbf{h}_\perp = \mathbf{h} - \mathbf{h}_\parallel$ be a part of $\mathbf{h}$ that is perpendicular to $\mathcal{U}$. It can be verified that $\mathbf{h}_\perp \in \mathcal{H}_L$ and for any $1 \leq i \leq m + u$, $\langle \mathbf{h}_\perp, U(i, \cdot) \rangle_L = 0$. For any $1 \leq i \leq m + u$ we have

$$
\begin{aligned}
h(i) &= \langle \mathbf{h}, U(i, \cdot) \rangle_L = \langle \sum_{j=1}^{m+u} \alpha_j U(j, \cdot), U(i, \cdot) \rangle_L + \langle \mathbf{h}_\perp, U(i, \cdot) \rangle_L \\
&= \sum_{j=1}^{m+u} \alpha_j \langle U(j, \cdot), U(i, \cdot) \rangle_L = \sum_{j=1}^{m+u} \alpha_j U(i, j) = h_\parallel(i) \;\; .
\end{aligned} \tag{72}
$$

The second equation in (72) holds by Lemma 13. As a consequence of (72), the empirical error (the second term in (71)) depends only on $\mathbf{h}_\parallel$. Furthermore,

$$
\mathbf{h}^T L \mathbf{h} = \langle \mathbf{h}, \mathbf{h} \rangle_L = \|\mathbf{h}\|_L^2 = \| \sum_{i=1}^{m+u} \alpha_i U(i, \cdot) \|_L^2 + \|\mathbf{h}_\perp\|_L^2 \geq \| \sum_{i=1}^{m+u} \alpha_i U(i, \cdot) \|_L^2 \;\; .
$$

Therefore, for an $\mathbf{h}^* \in \mathcal{H}$ that minimizes (71), $\mathbf{h}_\perp^* = 0$ and $\mathbf{h}^* = \mathbf{h}_\parallel^* = \sum_{i=1}^{m+u} \alpha_i U(i, \cdot) = U\boldsymbol{\alpha}$. $\square$

## Appendix H. Proof of Lemma 9

Let $L_N \triangleq I - L = I - D^{-1/2} W D^{-1/2}$ be a normalized Laplacian of $W$. The eigenvalues $\{\lambda_i'\}_{i=1}^{m+u}$ of $L_N$ are non-negative and the smallest eigenvalue of $L_N$, denoted here by $\lambda_{\min}'$, is zero (Chung, 1997). The eigenvalues of the matrix $I - \beta L = (1 - \beta)I + \beta L_N$ are $\{1 - \beta + \beta\lambda_i'\}_{i=1}^{m+u}$. Since $0 < \beta < 1$, all the eigenvalues of $I - \beta L$ are strictly positive. Hence the matrix $I - \beta L$ is invertible and its eigenvalues are $\left\{\frac{1}{1-\beta+\beta\lambda_i'}\right\}_{i=1}^{m+u}$. Finally, the eigenvalues of the matrix $U$ are $\left\{\frac{1-\beta}{1-\beta+\beta\lambda_i'}\right\}_{i=1}^{m+u}$. Since $\lambda_{\min}' = 0$, the largest eigenvalue of $U$ is 1. Since all eigenvalues of $L_N$ are non-negative, we have that $\lambda_{\min} > 0$.

## Appendix I. Proofs from Section 7

**Proof of Corollary 2:** Let $\{A_i\}_{i=1}^\infty$ and $\{p_i\}_{i=1}^\infty$ be a set of positive numbers such that $\sum_{i=1}^\infty p_i \leq 1$. By the weighted union bound argument we have from (39) that with probability of at least $1 - \delta$ over the training/test set partitions, for all $A_i$ and $\mathbf{q} \in \Omega_{g, A_i}$,

$$\mathcal{L}_u(\widetilde{\mathbf{h}}_\mathbf{q}) \leq \widehat{\mathcal{L}}_m^\gamma(\widetilde{\mathbf{h}}_\mathbf{q}) + \frac{R_{m+u}(\widetilde{\mathcal{B}}_{g, A_i})}{\gamma} + c_0 Q \sqrt{\min(m, u)} + \sqrt{\frac{S}{2} Q \ln \frac{1}{p_i \delta}} \ . \tag{73}$$

We set $A_i \triangleq g_0 s^i$ and $p_i \triangleq \frac{1}{i(i+1)}$. It can be verified that $\sum_{i=1}^\infty p_i \leq 1$. For each $\mathbf{q}$ let $i_\mathbf{q}$ be the smallest index for which $A_{i_\mathbf{q}} \geq g(\mathbf{q})$. We have two cases:

**Case 1** $i_\mathbf{q} = 1$. In this case $i_\mathbf{q} = \log_s(\tilde{g}(\mathbf{q})/g_0) = 1$.

**Case 2** $i_\mathbf{q} \geq 2$. In this case $A_{i_\mathbf{q}-1} = g_0 s^{i_\mathbf{q}-1} < g(\mathbf{q}) \leq \tilde{g}(\mathbf{q})s^{-1}$, and therefore, $i_\mathbf{q} \leq \log_s(\tilde{g}(\mathbf{q})/g_0)$.

Thus we always have that $i_\mathbf{q} \leq \log_s(\tilde{g}(\mathbf{q})/g_0)$. It follows from the definition of $A_{i_\mathbf{q}}$ and $\tilde{g}(\mathbf{q})$ that $A_{i_\mathbf{q}} \leq \tilde{g}(\mathbf{q})$. We have that $\ln(1/p_{i_\mathbf{q}}) \leq 2\ln(i_\mathbf{q} + 1) \leq 2\ln\log_s(s\tilde{g}(\mathbf{q})/g_0)$. Substituting these bounds into (73) and taking into account the monotonicity of $R_{m+u}(\widetilde{\mathcal{B}}_{g, A_i})$ (in $A_i$), we have that with probability of at least $1 - \delta$, for all $\mathbf{q}$, the bound (40) holds. $\square$

**Proof of Theorem 4:** We require several definitions and facts from the convex analysis (Rockafellar, 1970). For any function $f : \mathbb{R}^n \to \mathbb{R}$ the *conjugate function* $f^* : \mathbb{R}^n \to \mathbb{R}$ is defined as $f^*(\mathbf{z}) = \sup_{\mathbf{x} \in \mathbb{R}^n} (\langle \mathbf{z}, \mathbf{x} \rangle - f(\mathbf{x}))$. The domain of $f^*$ consists of all values of $\mathbf{z}$ for which the value of the supremum is finite. A consequence of the definition of $f^*$ is the so-called *Fenchel inequality*:

$$\langle \mathbf{x}, \mathbf{z} \rangle \leq f(\mathbf{x}) + f^*(\mathbf{z}) \ . \tag{74}$$

It can be verified that the conjugate function of $g(\mathbf{q}) = D(\mathbf{q}\|\mathbf{p})$ is $g^*(\mathbf{z}) = \ln \sum_{j=1}^{|\mathcal{B}|} p_j e^{z_j}$. Let $\widetilde{\mathbf{h}}(i) \triangleq (h_1(i), \ldots, h_{|\mathcal{B}|}(i))$. In the derivation that follows we use the following inequality

(Hoeffding, 1963): if $X$ is a random variable such that $a \leq X \leq b$ and $c$ is a constant, then

$$\mathbf{E}_X \exp(cX) \leq \exp\left(\frac{c^2(b-a)^2}{8}\right) \ . \tag{75}$$

For any $\lambda > 0$ we have,

$$
\begin{aligned}
R_{m+u}(\widetilde{\mathcal{B}}_{g,A}) &= Q\mathbf{E}_{\boldsymbol{\sigma}} \sup_{\mathbf{q} \in \Omega_{g,A}} \langle \boldsymbol{\sigma}, \widetilde{\mathbf{h}}_{\mathbf{q}} \rangle = Q\mathbf{E}_{\boldsymbol{\sigma}} \sup_{\mathbf{q} \in \Omega_{g,A}} \left\langle \mathbf{q}, \sum_{i=1}^{m+u} \sigma_i \widetilde{\mathbf{h}}(i) \right\rangle \\
&= \frac{Q}{\lambda} \mathbf{E}_{\boldsymbol{\sigma}} \sup_{\mathbf{q} \in \Omega_{g,A}} \left\langle \mathbf{q}, \lambda \sum_{i=1}^{m+u} \sigma_i \widetilde{\mathbf{h}}(i) \right\rangle \\
&\leq \frac{Q}{\lambda} \left( \sup_{\mathbf{q} \in \Omega_{g,A}} g(\mathbf{q}) + \mathbf{E}_{\boldsymbol{\sigma}} g^* \left( \lambda \sum_{i=1}^{m+u} \sigma_i \widetilde{\mathbf{h}}(i) \right) \right) \tag{76} \\
&\leq \frac{Q}{\lambda} \left( A + \mathbf{E}_{\boldsymbol{\sigma}} \ln \sum_{j=1}^{|\mathcal{B}|} p_j \exp\left[ \lambda \sum_{i=1}^{m+u} \sigma_i \mathbf{h}_j(i) \right] \right) \tag{77} \\
&\leq \frac{Q}{\lambda} \left( A + \sup_{\mathbf{h} \in \mathcal{B}} \mathbf{E}_{\boldsymbol{\sigma}} \ln \exp\left[ \lambda \sum_{i=1}^{m+u} \sigma_i \mathbf{h}(i) \right] \right) \\
&\leq \frac{Q}{\lambda} \left( A + \sup_{\mathbf{h} \in \mathcal{B}} \ln \mathbf{E}_{\boldsymbol{\sigma}} \exp\left[ \lambda \sum_{i=1}^{m+u} \sigma_i \mathbf{h}(i) \right] \right) \tag{78} \\
&\leq \frac{Q}{\lambda} \left( A + \sup_{\mathbf{h} \in \mathcal{B}} \ln \exp\left[ \frac{\lambda^2}{2} \sum_{i=1}^{m+u} \mathbf{h}(i)^2 \right] \right) \tag{79} \\
&= Q \left( \frac{A}{\lambda} + \frac{\lambda}{2} \sup_{\mathbf{h} \in \mathcal{B}} \|\mathbf{h}\|_2^2 \right) \ . \tag{80}
\end{aligned}
$$

Inequality (76) is obtained by applying (74) with $f \triangleq g$ and $f^* \triangleq g^*$. Inequality (77) follows from the definition of $g$ and $g^*$. Inequality (78) is obtained by an application of the Jensen inequality and inequality (79) is obtained by applying $m + u$ times (75). By minimizing (80) w.r.t. $\lambda$ we obtain

$$R_{m+u}(\widetilde{\mathcal{B}}_{g,A}) \leq Q\sqrt{2A \sup_{\mathbf{h} \in \mathcal{B}} \|\mathbf{h}\|_2^2} \ .$$

Substituting this bound into (39) we get that for any fixed $A$, with probability at least $1-\delta$, for all $\mathbf{q} \in \mathcal{B}_{g,A}$

$$\mathcal{L}_u(\widetilde{\mathbf{h}}_{\mathbf{q}}) \leq \widehat{\mathcal{L}}_m^\gamma(\widetilde{\mathbf{h}}_{\mathbf{q}}) + \frac{Q}{\gamma}\sqrt{2A \sup_{\mathbf{h} \in \mathcal{B}} \|\mathbf{h}\|_2^2} + c_0 Q\sqrt{\min(m,u)} + \sqrt{\frac{S}{2}Q \ln\frac{1}{\delta}}.$$

Finally, by applying the weighted union bound technique, as in the proof of Corollary 2, we obtain the statement of the theorem. $\qquad\square$